%% file: _main.tex
\documentclass[authoryear,preprint,review,12pt]{elsarticle}

\usepackage{amssymb}

\usepackage{lineno}
\usepackage{linguex}
\usepackage{graphicx}
\usepackage{amsmath}
\usepackage{makecell}
\usepackage{longtable}
\usepackage[multiple]{footmisc}

\journal{Cognition}

\begin{document}
\renewcommand{\firstrefdash}{}

\begin{frontmatter}

\include{titlepage}

\begin{abstract}
We computationally implement and experimentally test the behavioral predictions of a dynamic neural model of lexical meaning in the framework of Dynamic Field Theory. 
We demonstrate the architecture and behavior of the model using as a test case the English lexical item \textit{have}, focusing on its polysemous use.
In the model, \textit{have} maps to a semantic space defined by two independently motivated continuous conceptual dimensions, connectedness and control asymmetry. 
The mapping is modeled as coupling between a neural node representing the lexical item and neural fields representing the conceptual dimensions.
While lexical \textit{knowledge} is modeled as a stable coupling pattern, real-time lexical meaning \textit{retrieval} is modeled as the motion of neural activation patterns between transiently stable states corresponding to semantic interpretations or readings.
Model simulations capture two previously reported empirical observations: (1) contextual modulation of lexical semantic interpretation, and (2) individual variation in the magnitude of this modulation.
Simulations also generate a novel prediction that the by-trial relationship between sentence reading time and acceptability should be contextually modulated.
An experiment combining self-paced reading and acceptability judgments replicates previous results and partially bears out the model's novel prediction.
Altogether, results support a novel perspective on lexical polysemy: that the many related meanings of a word are not categorically distinct representations; rather, they are transiently stable neural activation states that arise from the nonlinear dynamics of neural populations governing interpretation on continuous semantic dimensions.
Our model offers important advantages over related models in the dynamical systems framework, as well as models based on Bayesian inference.
\end{abstract}

\begin{highlights}
\item We propose a model of lexical meaning in the framework of Dynamic Field Theory (DFT)
\item Model simulations capture known sentence comprehension effects and generate a novel prediction
\item Sentence reading experiment replicates previous results and confirms new model prediction
\item We argue that polysemy arises from nonlinear neural dynamics on continuous dimensions
\end{highlights}


\end{frontmatter}

\tableofcontents
\newpage



\section{Introduction}
\label{sec:intro}

In language comprehension, linguistic forms evoke interpretations of meaning.
A basic linguistic unit is the lexical item, a systematic relation between phonetic/phonological, morphological, syntactic, and semantic information \citep[e.g.,][]{anderson_-morphous_1992, chomsky_aspects_1965, jackendoff_morphological_1975}. 
It is well-attested that the specific interpretation or reading evoked by a lexical item can vary depending on context.  
\textit{Polysemy} refers to the situation whereby a lexical item offers more than one interpretation yet those possibilities are conceptually related. 
For example, the lexical item \textit{book} can be described as polysemous since its possible readings include (at least) both a physical object (e.g., \textit{heavy \underline{book}}), and information represented by the object (e.g., \textit{enjoyable \underline{book}}) \citep[e.g.,][]{brugman_story_1988,deane_polysemy_1988,lakoff_women_1990,pustejovsky_generative_1995,vicente_polysemy_2018}.\footnote{Polysemy contrasts with \textit{homophony}, the situation where the possible readings appear semantically unrelated, e.g., \textit{river \underline{bank}} vs. \textit{savings \underline{bank}}.}
Here we investigate the semantic basis of the constrained variability observed in lexical polysemy.
To this end, we examine a possible source of this variability, propose a neuro-computational implementation in the framework of Dynamic Field Theory \citep{schoner_dynamic_2016}, and investigate its behavioral predictions during sentence comprehension. 
Under our proposal, the many-to-one relation between meaning and form which is salient in cases of lexical polysemy represents the norm rather than the exception, with differences primarily in the magnitude of variability and the relatedness of the possible readings.\footnote{This work is situated in an analytical approach to lexical polysemy whereby meaning variability depends on contextual satisfaction of \textit{discrete} meaning representations that are, as it were, sequestered in the mental lexicon, and connected to a unified bundle of morphosyntactic and morphophonological properties \citep[e.g.,][]{deo_quantification_2011, jackendoff_morphological_1975, jackendoff_architecture_1997, pinango_reanalyzing_2016, pustejovsky_generative_1995}. Our proposal here shares with that traditional approach the assumption that the meaning generation process involved in polysemy is grounded in the combinatorial properties of the semantic system. It departs from it by placing the burden of meaning generation, including polysemy, on the interaction between the conceptual context of a lexical item and a \textit{continuous} meaning space. The result of this interaction is what gives the language user the experience of meaning discreteness: what we refer to as a lexical meaning, which is nonetheless never isolated from the larger continuous meaning space. On this view, meaning discreteness is not an input to semantic composition, contrary to what is traditionally assumed, but an outcome of the compositional process itself \citep{pinango_solving_2023}. To our knowledge, this is the first formal neuro-computational exploration of this idea.}

\subsection{English `have'}
\label{sec:have}
Our test case is the English lexical item \textit{have}. 
\textit{have} is typically interpreted as a relation of possession between the referents of its arguments, as in \ref{ex:possanim} where the possession is alienable, or as in \ref{ex:possinanim} where the possession is inalienable.

\ex.\label{ex:possanim} The oak tree has colorful decorations. (\textit{alienable possession})

\ex.\label{ex:possinanim} The oak tree has a healthy trunk. (\textit{inalienable possession})

When people are presented with less frequent argument combinations, as in \ref{ex:badPoss} below, two types of responses are reported:
(a) the inanimate subject ``the oak tree" leads to an inalienable possession reading, which conflicts with the fact that the object ``the motorcycle" does not plausibly enter into a part-whole relation with ``the oak tree", leading to a decrease in acceptability; or
(b) the referent of the subject is anthropomorphized, e.g., ``the person dressed as a tree...", which maintains an alienable possession interpretation \citep{zhang_real-time_2018,zhang_linguistic_2021}.
The facts presented so far are consistent with the generalization that \textit{have} primarily evokes possession readings.

\ex.\label{ex:badPoss} \#The oak tree has a motorcycle.\footnote{``\#" signifies that the sentence, while grammatical, is judged less felicitous without additional context.}

But the polysemy reportoire of \textit{have} is larger than these two readings.
\textit{have} can also evoke an interpretation of coincidental location. 
This would render \ref{ex:badPoss} above with an interpretation that the motorcycle is in coincidental spatial proximity to the oak tree (e.g., next to, under, below, above), an interpretation that we will refer to as \textit{adjacency} \citep[e.g.,][]{myler_building_2016,zhang_word-meaning_2022}. 
The tendency for \textit{have} to evoke a possession reading is very strong, but not indefeasible. 
One standard way to bring out an adjacency reading from \textit{have} is to add a prepositional phrase which makes explicit the spatial relation, as in \ref{ex:loc}. 
Not surprisingly, inclusion of the locative modifier ``next to it" is reported to improve acceptability ratings relative to sentences like \ref{ex:badPoss} \citep{zhang_real-time_2018,zhang_word-meaning_2022}.

\ex.\label{ex:loc} The oak tree has a motorcycle \underline{next to it}.

Inclusion of an explicit locative modifier is not the only way to make an adjacency reading salient, however. 
A bias for an adjacency interpretation can also be induced by the preceding context. 
Specifically, when a preceding sentence evokes an adjacency reading, as in \ref{ex:loc}, then sentences like \ref{ex:badPoss} are more likely to evoke an adjacency reading as well as receive higher acceptability ratings \citep{zhang_real-time_2018,zhang_word-meaning_2022}, as seen in \ref{ex:primeLoc}.

\ex.\label{ex:primeLoc} \underline{The pine tree has a car next to it} and the oak tree has a motorcycle.

This linguistic and behavioral pattern suggests a unified polysemy-based analysis of English \textit{have} involving readings ranging from a purely coincidental spatial relation to an alienable possession relation to an inalienable possession relation \citep{zhang_linguistic_2021}.
Crucially, these readings are not disconnected.
Evidence suggests that the distinction between adjacency, alienable possession, and inalienable possession is a matter of degree, not category, not only between these three reading types but also within them \citep[e.g.,][]{deo_diachronic_2015, zhang_linguistic_2021, zhang_word-meaning_2022}. 
In the case of coincidentality in spatial location, for example, the situation conveyed by ``the tree has a motorcycle" is more coincidental than ``the tree has a bench", and that, in turn, is more coincidental than ``the tree has a swing", or ``the tree has a nest". 
So, with each situation we move towards less coincidentality and, interestingly, towards greater control asymmetry between arguments such that the first argument is more and more perceived as ``possessing" the second.  

Similar gradability is observable within the alienable and inalienable possession spaces. 
For alienable possession, ``the woman has a car" can refer to a variety of related situations distinguishable by degree of ownership: the woman has a car because she stole it, because she borrowed it, rented it, bought it. 
Only in the last instance would she ``own" the car, yet in all instances the relation between the woman and the car---one of asymmetry in control---renders it one of possession. 
Finally, for the inalienable possession space, the possibilities range from the peripherally inalienable ``the woman has hair" to the more inalienable ``the woman has a liver", to the completely inalienable ``the woman has a body". 
Notably, while there is no coincidentality in these examples, making it more like the alienable possession space, the control relation between arguments—e.g., woman to her hair, woman to her liver, woman to her body—has become less and less asymmetrical, mirroring the relation in the coincidental location space  \citep{koch_location_2012,pinango_solving_2023,zhang_linguistic_2021}.

We conclude then that the lexical meaning of \textit{have}, while constrained, gives rise to gradable readings. 
As a result, we reason that such meaning is best characterized as a continuous space within which specific readings can obtain. 
This conclusion in turn raises the question of what the properties of such a continuous space are.
We address this question below.

\subsection{Continuous conceptual space for lexical meaning}
\label{sec:continuous_space}
It has been proposed that the readings of \textit{have} are supported by a conceptual space organized in terms of two continuous cognitive dimensions or metrics: \textit{control asymmetry} and \textit{connectedness} \citep[see also \citet{gardenfors_conceptual_2000,gardenfors_geometry_2014} for the notion of continuous conceptual dimensions and their relation to linguistic semantics]{pinango_concept_2019,pinango_solving_2023,zhang_linguistic_2021}.\footnote{In linguistics, the notion that continuous dimensions underlie apparently categorical distinctions has also been invoked in the domain of phonology \citep[e.g.,][]{browman_articulatory_1989,pierrehumbert_conceptual_2011,stevens_quantal_1989}. At the level of neural encoding, it is uncontroversial that dimensions underlying perception \citep[e.g.,][]{dyballa_population_2024,hubel_receptive_1959, leonard_large-scale_2024} and movement \cite[e.g.,][]{bouchard_functional_2013,chartier_encoding_2018,georgopoulos_neuronal_1986} are continuous.}
\textit{Control asymmetry} measures the degree to which two individuals in a situation differ in how much one controls the other. 
Control asymmetry underpins perception of causality and causal chains. 
In doing so, it gives rise to intuitions about intentionality and agency \citep[e.g.,][]{carey_origin_2009,croft_verbs_2012,talmy_force_1988}.
It has also been shown to constrain language creation in pidgins and creoles \citep[e.g.,][]{klein_utterance_1992}.
An evaluation of high control asymmetry between two participants means that one of the participants can be construed as a controller in the situation, and not the other way around. 
A low control asymmetry evaluation means that neither of the participants is construable as controller.\footnote{In the example sentences so far, as well as in the stimulus sentences in the experiment described below, both arguments of \textit{have} denote inanimate entities. Even though ``control" normally evokes animacy, it is not limited to it. Control asymmetry exists between inanimate entities whenever the state of one can be seen as depending on the other, and not vice versa; or when one can exert influence on the other, and not vice versa. In \ref{ex:possanim}, for example, the tree exerts greater influence on the decorations than vice versa, since the state of the decorations (in particular, the spatial position) depends more on the tree (since they are hanging on it) than vice versa, as the state of the tree would remain unchanged if the decorations were removed.} 

\textit{Connectedness} measures the degree to which two individuals in a situation are functionally or structurally related to one another. 
It is from connectedness that intuitions emerge about coherence in the world, e.g., object individuation and part-whole relations \citep[e.g.,][]{carey_origin_2009, gopnik_theory_2004, krojgaard_review_2004}.
Relations of connectedness are built along independently motivated conceptual dimensions—spatial, temporal, informational, or functional \citep{pinango_reanalyzing_2016}.

\begin{figure}[h]
    \centering
    \includegraphics[width=0.9\linewidth]{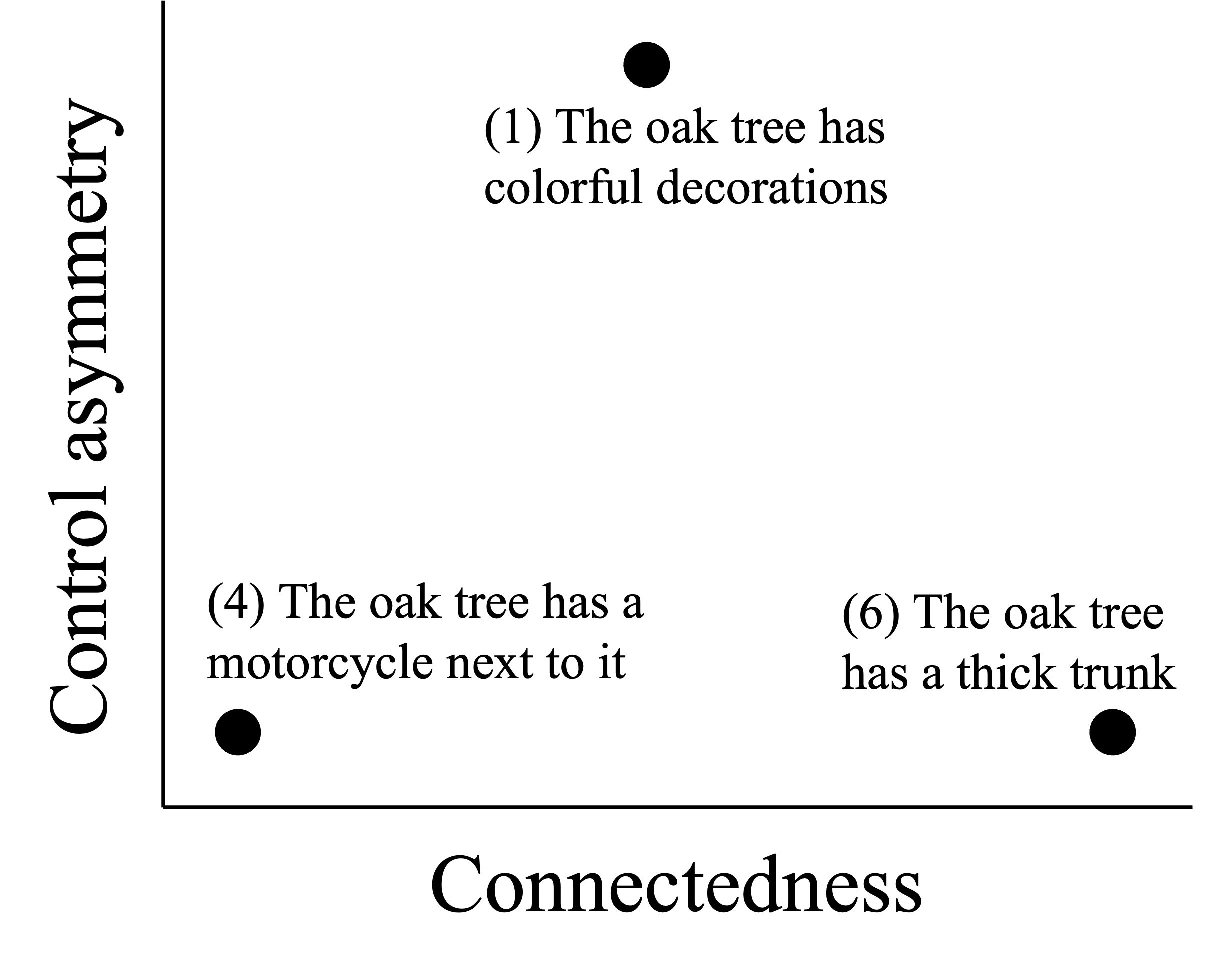}
    \caption{Three readings of \textit{have} plotted in continuous 2D semantic space. Numbers correspond to example sentence labels in the text.}
    \label{fig:MDS_sparse}
\end{figure}

In Figure \ref{fig:MDS_sparse}, three possible readings of \textit{have} are plotted in the space spanned by the two dimensions of connectedness (horizontal axis) and control asymmetry (vertical axis).
A coincidental adjacency reading, as in \ref{ex:loc}, corresponds to low connectedness and low control asymmetry, and an alienable possession reading, as in \ref{ex:possanim}, corresponds to higher connectedness and higher control asymmetry.
An inalienable possession reading, as in \ref{ex:partWhole}, corresponds to high connectedness and low control asymmetry.

\ex.\label{ex:partWhole} The oak tree has a thick trunk.

\begin{figure}[h]
    \centering
    \includegraphics[width=0.9\linewidth]{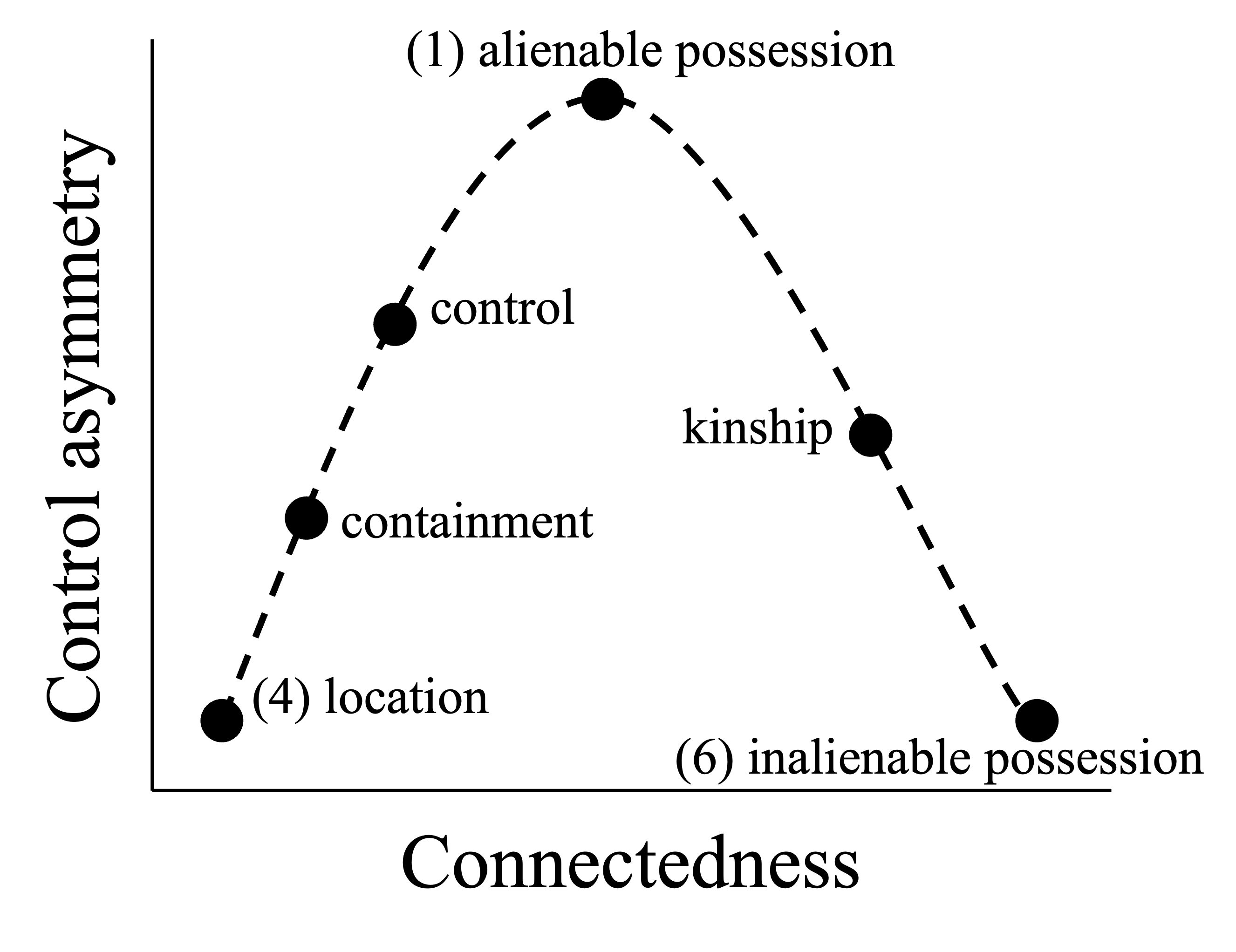}
    \caption{Six readings of English \textit{have}. The parabola indicates hypothesized constraints on the relationship between the two semantic dimensions. Adapted from \citet{pinango_concept_2019}.}
    \label{fig:MDS}
\end{figure}

As seen in Figure \ref{fig:MDS}, when more possible readings of \textit{have} are plotted (including, in addition to the examples given so far, at least containment, control, and kinship), they tend to fall on a downward parabola.
This pattern indicates dependencies between the two dimensions which constrain object perception and its linguistic encoding.
We return to this point in the context of our model in Section \ref{sec:model}.

In addition to the robust observation of gradability between readings of \textit{have}, empirical support for a model of lexical meaning based on interpretable, continuous dimensions also comes from trajectories of meaning change over time.
Over an approximately 200-year period, the postposition \textit{kade} in the Indo-Aryan language Marathi has shifted gradually from primarily an adjacency reading, to primarily an alienable possession reading, to primarily an inalienable possession reading \citep{deo_diachronic_2015,zhang_linguistic_2021}. 
This trajectory essentially traces the downward parabola shape in Figure \ref{fig:MDS} from left to right. 
This supports the notion that the potential readings of \textit{have} are organized along these continuous dimensions, rather than forming a countable set of discrete meanings.\footnote{An anonymous reviewer notes that the adjacency and possession readings that we have focused on here hardly exhaust the polysemic repertoire of \textit{have}. Cases like \textit{they have a miscarriage/a conniption/a seizure} or cases like \textit{have mercy/pity/compassion} (in the imperative form), or \textit{they have a banana/a bite/dinner} are also available readings, yet at first glance they don’t fall within either of those groupings. Full discussion of these cases is outside of the scope of the paper. However, we want to provide a sense of how they are captured within the space. The first three cases represent inalienable possession readings (high connectedness, low control asymmetry). \textit{have a miscarriage} is a variant in that the eventuality of \textit{miscarriage} demands a shift in connectedness (from high to mid) implementing the process of alienation between the mother and the fetus. In the second set, \textit{mercy}, \textit{pity} and \textit{compassion} are emotional manifestations of a mind, indicating high connectedness and normally low control asymmetry. Yet, the use in the imperative construction conveys the possibility of control asymmetry at some level. Finally, the consumption readings (\textit{have a banana/bite/dinner}) involve a process (the act of ingesting) that results in a state: the point \textit{after} ingestion has taken place, resulting in the theme changing location. As in the case of \textit{miscarriage}, these readings demand a shift, a move from mid to higher connectedness and to lower control asymmetry (closer to the part-whole/inalienable possession space), revealing the change in relation between the two participants that ingestion creates. This is because whatever is ingested becomes part of the organism that ingested it and ultimately can have the power to “control” such organism. This is evident in cases like, e.g., \textit{have the hemlock} where the controllee once ingested becomes the controller of sorts of the host organism.}\footnote{Some aspectual uses of \textit{have} appear to be outside the uses alluded to here. Indeed, the auxiliary uses represent a further grammaticalization of \textit{have} brought about by morphosyntactic reanalysis during Old English of the object and the past participle; a reanalysis which was made possible by the freer word order of the language at that time \cite[e.g.,][]{michaelis_toward_1993,pancheva_perfect_2003}. Specifically, the English \textit{have}-perfect appears to have emerged from a reanalysis of the complex transitive use, indicating possession---as in “I [have$_v$ [[a house$_{obj}$] [broken$_{adj}$]]”---to a transitive use where the possession reading is no longer available, as in “I [have [broken [a house]]]”. This left \textit{have} to take on an additional semantically distinct use, i.e., end sub-interval of the event, thus creating a novel lexical item dedicated to this semantic use but only when appearing in this “split predication” configuration (ibid.).} 

\subsection{Real-time comprehension of \textnormal{have}}
\label{sec:empirical_support}
A range of real-time comprehension evidence is consistent with the hypothesis that the lexical meaning of \textit{have} is defined on these two continuous dimensions.
Such evidence comes from acceptability judgments, self-paced reading, and electroencephalography (EEG) experiments investigating the availability of the coincidental adjacency reading associated with \textit{have} as a result of preceding context. 
\citet{zhang_real-time_2018} administered an acceptability judgment task with target sentences consisting of \textit{have} and two unrelated, inanimate arguments, as in \ref{ex:badPoss}.
Each target sentence was preceded by a context sentence, also consisting of \textit{have} and two inanimate arguments.
The crucial manipulation was the meaning of the context sentence: adjacency, as in \ref{ex:loc}, or inalienable possession, as in \ref{ex:inalienablePoss}.

\ex.\label{ex:inalienablePoss} The pine tree has big branches.

When preceded by a context sentence designed to evoke an adjacency reading, the target sentence received higher acceptability ratings, relative to when the same target sentence was preceded by a context sentence designed to evoke a possession reading.
This improvement in acceptability is attributed to a gradient shift in interpretation towards adjacency due to the influence of the context sentence.

Using a similar stimulus set as in the acceptability judgment task, \citet{zhang_real-time_2018} observed decreased reading times for target sentences preceded by an adjacency context sentence, relative to when they were preceded by a possession context sentence.
Moreover, EEG recordings indicated an N400 event-related potential (ERP) in the possession condition relative to the adjacency condition \citep{zhang_real-time_2018}.
These results are consistent with a decreased neurocognitive processing load in the adjacency condition, due to contextual facilitation of an adjacency reading of the target sentence.

Finally, the magnitude of contextual modulation of \textit{have} interpretation correlates at an individual level with an independent measure of context sensitivity, i.e., the Autism-Spectrum Quotient \citep[AQ:][]{baron-cohen_autism-spectrum_2001}.
Individuals with higher AQ scores (decreased context sensitivity) exhibit a decreased difference in acceptability between conditions \citep{zhang_word-meaning_2022}.
These results suggest that the influence of context on interpretation is a gradient rather than categorical phenomenon, and some of the variation in the magnitude of contextual influence can be explained by the AQ.
However, these results do not necessarily rule out a model based on discrete meaning representations. 
We discuss this alternative possibility below.

\subsection{A discrete alternative}
\label{sec:discrete_alternative}

One could propose that the meaning of \textit{have} is represented as a countable set of discrete meanings, including, at least, possession and adjacency.
Preceding context would bias the comprehender towards one of these discrete options, and when the target sentence violates the comprehender's expectation, surprisal would result in lowered acceptability, slower reading, and modulation of electrophysiological responses \citep[e.g.,][]{frazier_sentence_1987,levy_expectation-based_2008}.
Gradience in such measurements could result from randomly distributed noise around underlyingly discrete responses types.

Such a proposal is particularly appealing when only two possible readings are considered, as in the experiments described above, as well as in the experiment reported below.
However, the broader range of empirical evidence reviewed above supports the continuous account: the synchronic range of \textit{have} interpretation, the diachronic trajectory of change in the meaning of Marathi \textit{kade}, and evidence for continuous cognitive dimensions in other domains.
The continuous representational account also has two theoretical advantages over a discrete representational account: parsimony and explanatory power.
Under a discrete representational account, a large number of representations would have to be posited: one for each reading. 
The continuous account is more parsimonious because it includes fewer representations: only two (continuous) dimensions. 
It is also more explanatory because it explicitly relates the different readings to each other, via direction and distance in the continuous space. 
Ultimately, extending our approach to the larger variety of possible interpretations of \textit{have}--—as well as other cases of polysemy—--will be necessary to support the hypothesis of continuous representations for lexical meaning.
A main contribution of this paper is to offer a neural processing model of lexical meaning on continuous cognitive dimensions, which is a necessary step towards extending to other interpretations and other lexical items.
To situate our neural processing model, we turn now to Dynamic Field Theory \citep[DFT:][]{schoner_dynamic_2016}, a formal framework for understanding the neural activation dynamics underlying continuous cognitive dimensions.

\subsection{Dynamic Field Theory (DFT)}
\label{sec:DFT}
In this subsection, we give a brief overview of DFT; a more detailed description of our model is given in Section \ref{sec:model} and \ref{appendix:equations}.
In DFT, dimensions relevant for cognition are modeled as continuous parameters governed by the activity of populations of neurons.
The activity of a neural population over time is described using a differential equation defining a \textit{dynamic neural field} \citep[DNF:][]{amari_dynamics_1977}.
DNFs are characterized by \textit{point attractor dynamics}. 
This means that, at any given time, the activation pattern in a DNF is attracted to a particular state, i.e., the point attractor state. 
The location of the point attractor can change over time according to a variety of factors, e.g., inputs to the DNF from sensory surfaces or other DNFs. 
Since change within DNFs is characterized by point attractor dynamics, but the location of the point attractor changes over time, DNFs are characterized by an interplay between stability and flexibility.
Stability offers resistance to the ubiquitous influence of noise, and flexibility allows rapid change under changing cognitive and environmental conditions \citep[e.g.,][]{kelso_dynamic_1995}.
Usually, when a DNF is not receiving any input, the point attractor corresponds to a resting state.
When a DNF begins to be influenced by input, the point attractor might shift to an active state.
In particular, the dynamics of lateral interaction within DNFs allow the formation of ``peaks" of activation.
Depending on the cognitive dimension being represented by the DNF, an activation peak might correspond to a movement goal, a percept, or another kind of cognitive event.
Discontinuous shifts from inactive states to active states (and vice versa) exemplify the nonlinear dynamics of DNFs, whereby continuous change in one variable (e.g., input to a DNF) can result in sudden, discontinuous change in another variable (e.g., DNF activation). 
In the context of lexical meaning, nonlinear dynamics offer a way to reconcile intuitions of discreteness with evidence for underlying continuity.

\subsubsection{DFT and language}
\label{sec:DFT_language}
DFT originally developed in the context of motor control research, especially in the domains of eye movements \citep{kopecz_saccadic_1995} and arm movements \citep{erlhagen_dynamic_2002}.
It has been increasingly applied in other cognitive domains, including, recently, speech and language.
A number of DFT models of speech and language have focused on the neurocognitive representation of phonetic dimensions, e.g., voice onset time (VOT), and the location and degree of constrictions formed by the tongue.
In these models, activation peaks correspond to articulatory movement goals.
These models have offered novel explanations for a variety of empirical phenomena:
effects of auditory perception on verbal response times \citep{roon_perceiving_2016},
effects of lexical competitors on speech articulation in errors \citep{stern_dynamic_2022} and non-errors \citep{stern_neural_2023,stern_not_2023},
long-term phonological change \citep{gafos_dynamical_2009,kirkham_dynamic_2024,shaw_dynamic_2023},
and individual differences in phonological representations \citep{harper_individual_2021}.
Other DFT models of language have focused on the processing of words and phrases describing physical properties of objects like color, size, and spatial position \citep{bhat_word-object_2022,kati_interaction_2024,richter_neural_2021,sabinasz_neural_2023}.
DFT is particularly useful for unifying discrete and continuous linguistic representations, and synthesizing empirical results and theoretical insights from different subfields in the study of human language \citep{stern_dynamic_2025}.

\subsection{This paper}
\label{sec:this_paper}
In this study, we extend neural field dynamics to semantic dimensions underlying lexical polysemy.
The model we propose implements a mapping between the lexical item \textit{have} and the continuous semantic space schematized in Figures \ref{fig:MDS_sparse} and \ref{fig:MDS}.
In the model, peaks of neural activation correspond to semantic interpretations.
One contribution of the paper is to offer a neurocognitive process-based explanation of the behavioral results described above: contextual modulation of the timecourse and outcome of lexical interpretation, and individual variation in the magnitude of this modulation.
Our explanation captures the intuition that meaning is experienced as discrete (e.g., ``adjacency" or ``possession"), despite evidence for a continuous substrate. 
Such apparent discreteness, we propose, arises from nonlinear dynamics of continuous neurocognitive variables.
In addition to capturing existing results, simulations from the model also generate a novel empirical prediction, which we test with an experiment combining self-paced reading and acceptability judgments.
We intend our model to be general, using the English lexical item \textit{have} as a test case. 
We propose that, while individual lexical items vary in the relevant semantic dimensions, as well as in the details of their coupling to those dimensions, the basic architecture and mechanisms of our model do not vary.

The remainder of the paper is structured as follows. 
Section \ref{sec:model} describes the structure of the model and the results of the simulations. 
Section \ref{sec:experiment} describes the design and results of the behavioral experiment. 
Section \ref{sec:general_discussion} discusses theoretical implications of the study and new predictions that remain to be tested. 

\section{Dynamic neural model of lexical meaning}
\label{sec:model}

\subsection{Model structure}
\label{sec:structure}

The model consists of two dynamic neural fields (DNFs) and one dynamic node.\footnote{This section describes the qualitative structure and behavior of the model. Model equations with more detailed descriptions are given in \ref{appendix:equations}.}
One DNF governs interpretation on the semantic dimension of control asymmetry (CA), and the other DNF governs interpretation on the dimension of connectedness (conn). 
The node corresponds to the lexical item \textit{have}. 
Activation of the node follows linear point attractor dynamics, with the position of the point attractor in activation space determined by the sum of external input to the node, e.g. from perception or intention, and random noise.

Activation in each of the two DNFs also unfolds according to point attractor dynamics.
Each DNF receives inputs (e.g., from the \textit{have} node, or from the other DNF, as described below) represented as Gaussian distributions with amplitude $a$, position $p$, and width $w$.
Under the influence of these time-varying inputs, the position of the point attractor in each DNF transitions between transiently stable resting states and transiently stable activation peaks, which correspond to semantic interpretations (see Section \ref{sec:DFT} and \ref{appendix:equations}).
Crucially, we set the parameters of lateral interaction such that only a single activation peak can form at a time in each DNF, defining ``selection" dynamics.
The range of each DNF is set from 1 to 99, such that each neuron in each DNF is maximally sensitive (``tuned”) to a particular percentage of the maximum conceivable value of that dimension. 
For instance, when the neuron $x = 80$ in the conn DNF is active, this corresponds to an interpretation of “$80\%$ of maximum conceivable connectedness”. 

\begin{figure}[h]
    \centering
    \includegraphics[width=0.8\linewidth]{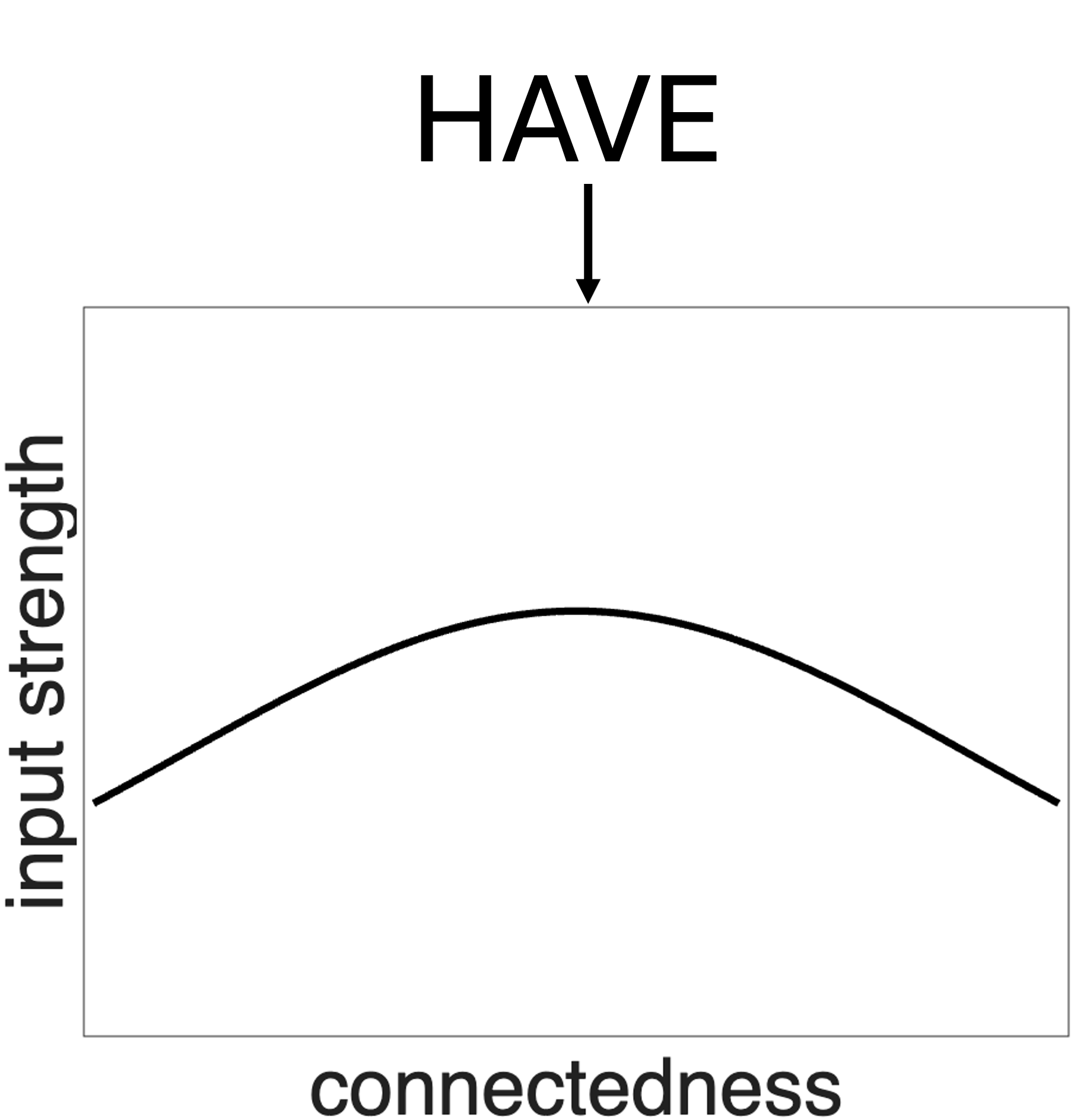}
    \caption{Distribution of input from the \textit{have} node to the conn DNF.}
    \label{fig:have_input}
\end{figure}

The \textit{have} node is coupled to the conn DNF with a wide distribution ($w_{\text{have}}$ = 40) positioned at the center of the field ($p_{\text{have}}$ = 50), as seen in Figure \ref{fig:have_input}. 
When the \textit{have} node becomes active, it sends input to the entire conn field, consistent with an analysis of polysemy for the lexical semantics of \textit{have}. 
However, the center of the conn field is favored, consistent with the fact that, all else equal, \textit{have} privileges an interpretation of alienable possession (intermediate connectedness), as described in Section \ref{sec:have}.

\begin{figure}[h]
    \centering
    \includegraphics[width=0.8\linewidth]{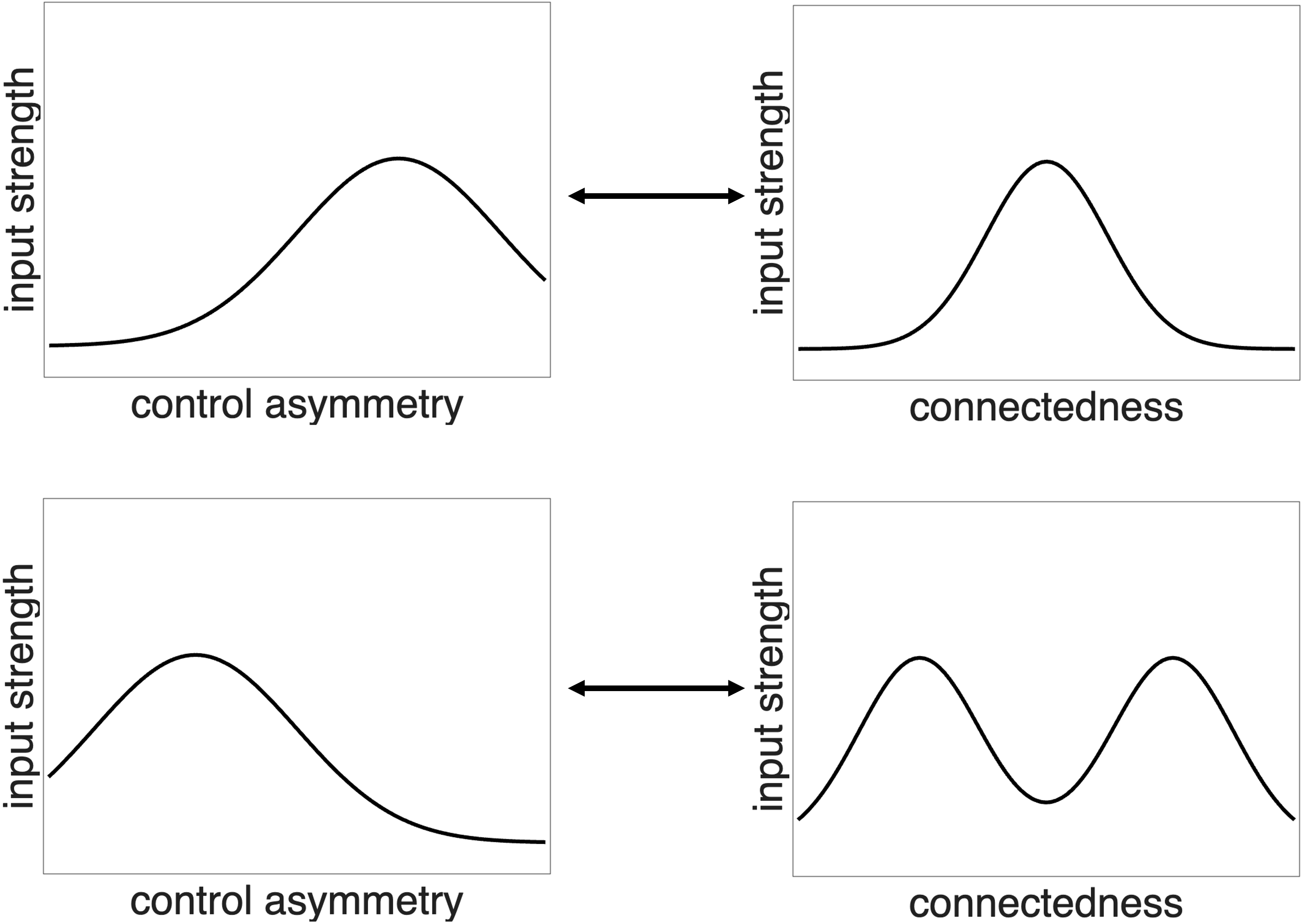}
    \caption{Distribution of input from each DNF to the other.}
    \label{fig:DNF_coupling}
\end{figure}

The two DNFs are also coupled to each other, such that active neurons in each DNF send input to the other DNF. 
Via this mechanism, patterns of activation in one DNF evoke corresponding patterns of activation in the other DNF. 
This mechanism implements the downward parabola pattern described in Section \ref{sec:intro} (Figure \ref{fig:MDS}). 
In particular, as seen in Figure \ref{fig:DNF_coupling}, activation consistent with high control asymmetry evokes activation consistent with intermediate connectedness (and vice versa; note the double-sided arrows), and activation consistent with low control asymmetry evokes activation consistent with both low connectedness and high connectedness (and vice versa). 

\subsection{Model simulations}
\label{sec:simulations}

In this section, we use the model to simulate interpretation of sentences containing \textit{have}.
The main purpose of the simulations is to investigate how context influences interpretation of anomalous sentences like \ref{ex:badPoss}, analogous to the experiment from \cite{zhang_real-time_2018}, described in Section \ref{sec:empirical_support}.
We present these simulation in Sections \ref{sec:example_context} and \ref{sec:1000_simulations}.
First, in Section \ref{sec:example_alienable}, we demonstrate how the model interprets a more canonical use of \textit{have}, as in \ref{ex:canonPoss}.

\ex.\label{ex:canonPoss} The professor has a motorcycle.

\subsubsection{Example simulation: alienable possession}
\label{sec:example_alienable}
\begin{figure}[h]
    \centering
    \includegraphics[width=1\linewidth]{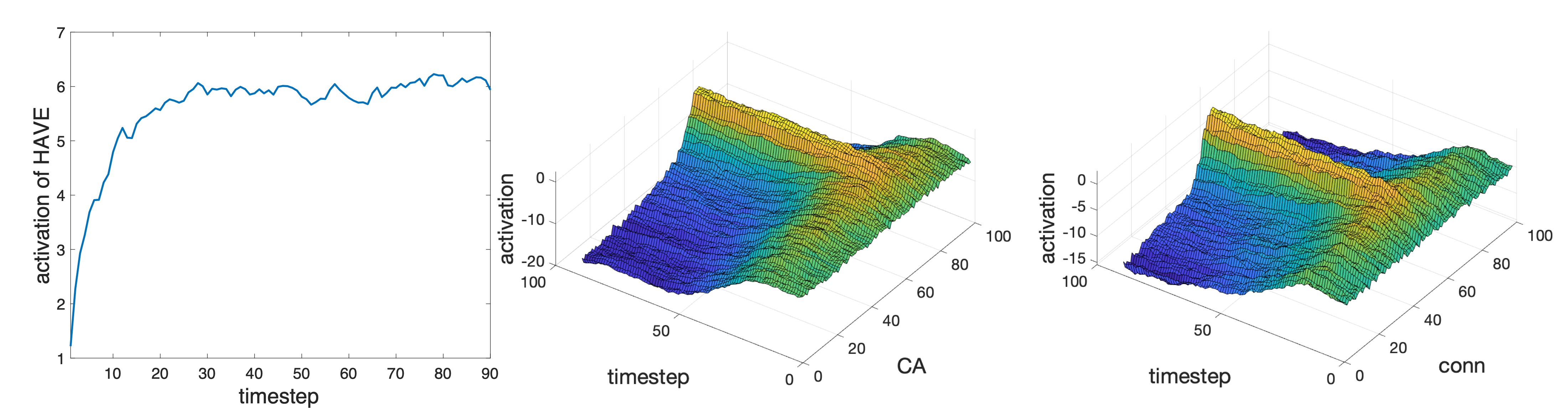}
    \caption{Activation history of the \textit{have} node (left), the CA field (center), and the conn field (right) under the influence of inputs corresponding to sentence \ref{ex:canonPoss}.}
    \label{fig:activation_history_canonical}
\end{figure}

To simulate interpretation of the sentence in \ref{ex:canonPoss}, the model receives three external inputs, all with amplitude $a = 6$: input to the \textit{have} node, input corresponding to high CA, and input corresponding to mid conn.
The input to the \textit{have} node corresponds to perception of the lexical item \textit{have}.
The other two inputs correspond to knowledge about the referents of the nouns serving as arguments of \textit{have}: \textit{professor} and \textit{motorcycle}.
We simulated model evolution under the influence of these three inputs for 90 timesteps.
Figure \ref{fig:activation_history_canonical} displays the results.

It can be seen that all three model components stabilize in active states shortly after the onset of the simulation.
Activation of the \textit{have} node is attracted to $u = 6$ because of the external input to the node.
The CA field forms an activation peak at the high end of the field because of external input centered at this location, as well as input from the conn field centered at this location (see Figure \ref{fig:DNF_coupling}). 
The conn field forms an activation peak in the middle of the field because of external input, input from the \textit{have} node, and input from the CA field.
This qualitative activation state will persist until the inputs change or are removed. 
In this example, the model stabilizes quickly, and predictably, because the various influences on activation---external inputs, node-field coupling, and field-field coupling---reinforce each other.

\subsubsection{Example simulations: contextual modulation}
\label{sec:example_context}

Next, we simulate sentences containing \textit{have} with two unrelated, inanimate arguments (as in sentence \ref{ex:badPoss}) in two contexts: following an ``adjacency" reading of \textit{have}, and following a ``possession" reading of \textit{have}.\footnote{We use the terms ``adjacency" and ``possession" as labels to differentiate the two context sentence types, with the understanding that, under our proposal, these labels correspond to regions within a continuous semantic space, rather than discrete representations.}
These simulations are analogous to the experiment in \cite{zhang_real-time_2018}, described in Section \ref{sec:empirical_support}.
Each simulation consists of three phases, summarized in Table \ref{tab:sim_cond}.
In \textbf{phase 1}, the model receives external inputs which drive interpretation of the context sentence. 
In the adjacency context (corresponding to sentence \ref{ex:loc}), these inputs correspond to low CA and low conn because of the meanings of the nouns and the locative prepositional phrase; the \textit{have} node also receives input. 
In the possession context (corresponding to sentence \ref{ex:inalienablePoss}), external inputs also excite the \textit{have} node and the low side of the CA field, but the conn field receives input corresponding to high connectedness, i.e., inalienable possession. 
Again, the latter two inputs come from the meanings of the nouns which are the arguments of \textit{have}.
\textbf{Phase 1} is the only phase which differs between conditions. 
In \textbf{phase 2}, all inputs are removed. 
This corresponds to the time between interpreting the context sentence and interpreting the target sentence. 
In \textbf{phase 3}, the model receives external inputs corresponding to the target sentence. 
These inputs are identical in both conditions: \textit{have} ($a = 6$), low CA ($a = 6$), and low conn ($a = 0.4$).
The amplitude of the input to the low end of the conn field is much weaker in the target sentence relative to the adjacency context sentence, because the target sentence lacks a locative prepositional phrase.
Nonetheless, the two unrelated, inanimate nouns in the target sentence are assumed to induce a weak bias for a low connectedness interpretation.
\textbf{Phases 1} and \textbf{3} each run for 90 timesteps, which was found to be enough time for an activation peak to stabilize in both DNFs. 
\textbf{Phase 2} runs for 20 timesteps, which was found to be enough time for both activation peaks to fall below the interaction threshold ($u = 0$), but not enough time for the fields to return fully to the resting level. 
Each simulation runs for a total of 200 timesteps.

\begin{table}
    \centering
    \begin{tabular}{p{3cm}|p{3cm}|p{3cm}|p{3cm}}
    \hline
         & \thead{\textbf{Phase 1:} \\ \textbf{context}} & \thead{\textbf{Phase 2:} \\ \textbf{no input}} & \thead{\textbf{Phase 3:} \\ \textbf{target}} \\
         \hline
      \thead{\textbf{adjacency} \\ \textbf{context}} & \makecell{low CA (6) \\ \textbf{low conn} (6) \\ \textit{have} (6)} & - & \makecell{low CA (6) \\ low conn (0.4) \\ \textit{have} (6)} \\
      \hline
      \thead{\textbf{possession} \\ \textbf{context}} & \makecell{low CA (6) \\ \textbf{high conn} (6) \\ \textit{have} (6)} & - & \makecell{low CA (6) \\ low conn (0.4) \\ \textit{have} (6)} \\
      \hline
    \end{tabular}
    \caption{Summary of external inputs $s_{\text{ext}}$ to each model component in each phase of each simulation. Numbers in parentheses indicate the amplitude $a$ of each input.}
    \label{tab:sim_cond}
\end{table}
 
\begin{figure}[h]
    \centering
    \includegraphics[width=1\linewidth]{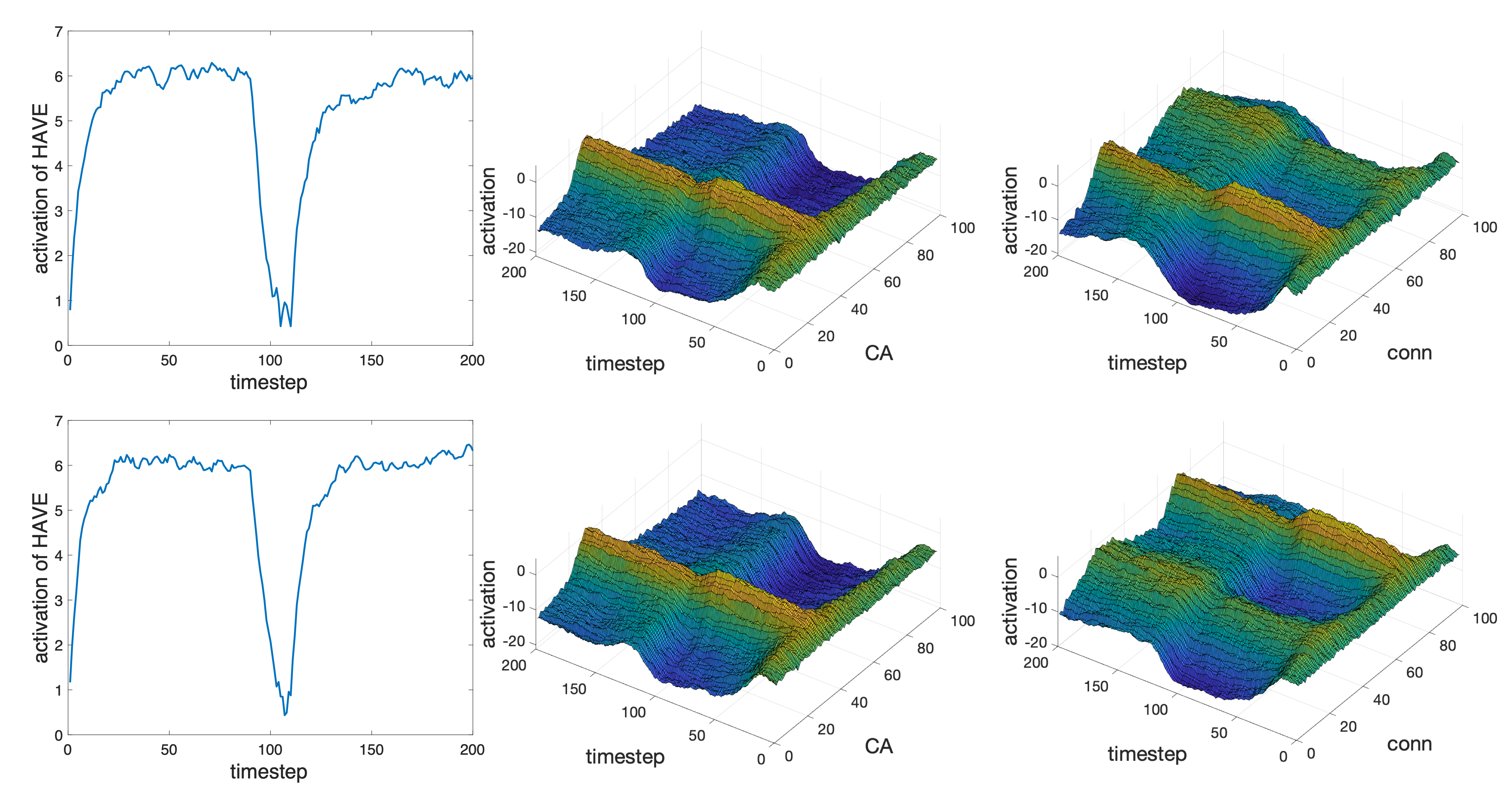}
    \caption{Activation history of the \textit{have} node (left), the CA field (center), and the conn field (right) in the adjacency context (top) and the possession context (bottom) for a single simulation.}
    \label{fig:activation_history}
\end{figure}

Examples of simulated interpretation in each context are displayed in Figure \ref{fig:activation_history}. 
In \textbf{phase 1} (up to timestep 90) the \textit{have} node becomes active, and begins to send input to the conn field (see Figure \ref{fig:have_input}). 
The CA field quickly forms a stabilized peak corresponding to an interpretation of low control asymmetry in both conditions. 
Due to field coupling, this CA peak sends input to the conn field on both the low and high ends (see Figure \ref{fig:DNF_coupling}). 
In the adjacency context, an activation peak corresponding to low conn stabilizes and suppresses the rest of the field via lateral inhibition. 
In the possession context, the winning activation peak is on the high side of the conn field. 
During \textbf{phase 2} (from timesteps 91–110), all activation decreases towards resting level. 
Activation of the \textit{have} node returns almost back to its resting level. 
In the fields, activation falls below the interaction threshold ($u = 0$), but does not return fully to the resting level. 
At timestep 111 (the beginning of \textbf{phase 3}), field activation is still highly non-uniform, reflecting residual activation from \textbf{phase 1}. 
In the CA field, another peak forms on the low side of the field in both conditions. 
In the conn field, the location of the peak differs by condition, despite the fact that the \textbf{phase 3} inputs are identical between conditions. 
In the adjacency context, \textbf{phase 3} (target sentence) processing stabilizes on another interpretation of low conn; in the possession context, the stabilized peak corresponds to an interpretation of high conn. 
In this way, the evolution of activation in \textbf{phase 3} is shaped by the preceding state of the system from \textbf{phase 1}.

\subsubsection{1000 simulations per condition}
\label{sec:1000_simulations}

\begin{figure}[h]
    \centering
    \includegraphics[width=0.7\linewidth]{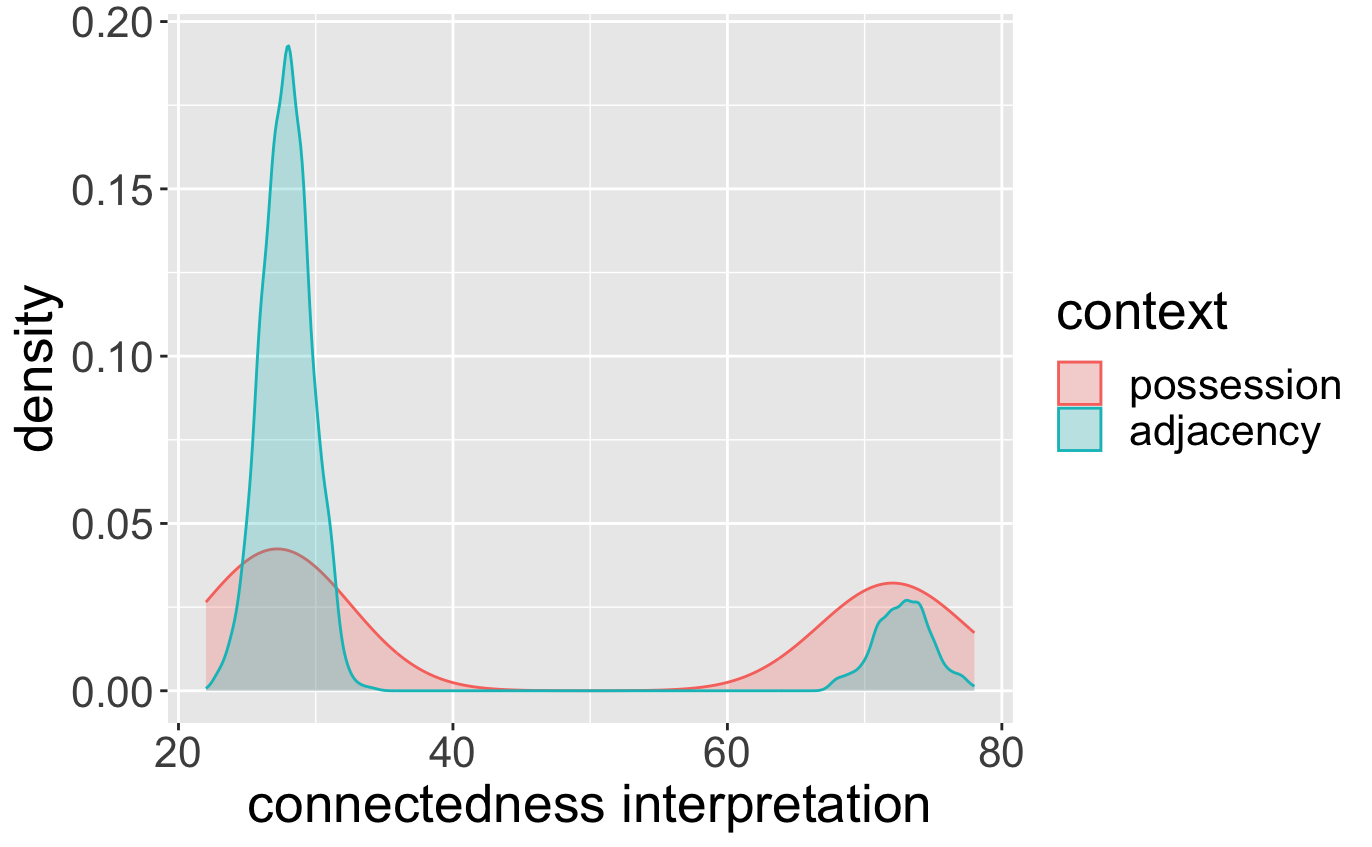}
    \caption{Distribution of activation peak location in the conn field at the end of each of 1000 simulations in each context.}
    \label{fig:peak_distribution}
\end{figure}

The examples in Figure \ref{fig:activation_history} were selected in order to demonstrate the effect of preceding context (\textbf{phase 1} evolution) on target sentence interpretation (\textbf{phase 3} evolution). 
However, not every simulated run of the model exhibits the same effect. 
The presence of noise in the model introduces a stochastic influence on the location of field stabilization in each simulation. 
Moreover, the weak low conn input in \textbf{phase 3} introduces an overall bias for low connectedness in both contexts.
In order to examine the robustness of the contextual modulation effect, we simulated 1000 instances of interpretation in each of the two contexts. 
As seen in Figure \ref{fig:peak_distribution}, there is a bimodal distribution of interpretations in both contexts. 
That is, for each context, an interpretation corresponding to either adjacency or possession for the target sentence was possible. 
In this way, the behavior of the model can be qualitatively described as corresponding to discrete outcomes, despite the underlyingly continuous state space (continuous features, continuous time, continuous activation).
The apparent discreteness of the interpretations of \textit{have} is consistent with the intuitions of experiment participants \citep{zhang_linguistic_2021} and linguistic researchers \citep{myler_building_2016}.
Importantly, this apparent discreteness emerges from the structure imposed by the coupling patterns (node-field and field-field), along with the nonlinear activation dynamics which drive selection of a particular field location on each simulation.
Nonetheless, there is gradient variability in the precise location of each activation peak, due to noise in each of the model components.

Importantly, the likelihood of each interpretation (low conn or high conn) was influenced by context. 
In the adjacency context, low conn (adjacency) interpretations were much more likely, due to the combined influence of the context and the weak bias for low conn coming from the nouns in \textbf{phase 3}.
In the possession context, high conn (possession) and low conn (adjacency) interpretations were approximately equally likely, due to the competing influences from the context (biased towards high conn) and from the bias for low conn coming from the nouns.
Thus, while context does not completely determine the course of field evolution, it exerts a strong enough influence to be observable over many simulations. 

\subsection{Simulated acceptability}
\label{sec:simulated_acceptability}
How do we relate these simulated results to the observed acceptability results described in Section \ref{sec:intro}? 
Modeling human acceptability judgments is far from trivial.
The process of making an acceptability judgment is influenced by a variety of linguistic and non-linguistic factors (for discussion of some of these issues, see \cite{cowart_experimental_1997,lau_measuring_2014,schutze_empirical_1996,sprouse_colorless_2018}).
Here, we propose a particular operationalization of acceptability that is intended to capture meaningful variation in the context of this complexity.
We propose that acceptability judgments are related to the \textit{distance between the activation peak in a neural field and the centroid of the distribution of input to the neural field}.
More specifically, we define acceptability as the reciprocal of one plus the distance between the location of the activation peak and the centroid of the input distribution, as in Eq. \ref{eq:acceptability}.
\begin{equation}\label{eq:acceptability}
\text{acceptability} = \frac{{1}}{{1 + \lvert x_{peak} - \bar{x}_{input} \lvert}}
\end{equation}
Thus, acceptability ranges from $\frac{{1}}{{F}}$ to $1$, where $F = $ the size of the field. 
Eq. \ref{eq:acceptability} indexes the degree to which an activation peak is consistent with ``expected" activation under this set of inputs, operationalized as the centroid of the input distribution. 
An actual activation peak at a given time can deviate from the ``expected" activation peak because of the influence of context and noise. 
The degree to which the activation peak deviates from the input centroid measures the degree to which an interpretation deviates from a canonical or expected interpretation.

According to this definition, a separate acceptability measure can be calculated for each field that (1) stabilizes in an “on” state (i.e., has an activation peak) and (2) receives input from the relevant lexical items (in this case, \textit{have} and its two arguments). 
In this study, we model two fields underlying interpretation of the target sentences: one representing connectedness (conn) and one representing control asymmetry (CA). 
However, under our simulation parameters (which are designed to mirror interpretation of the experimental sentences, described below), the only field in which the activation peak systematically diverges from the input distribution is the conn field. 
The location of the activation peak in the CA field does not vary systematically across trials. 
For this reason, in our calculation of acceptability, we focus only on the conn field. 
This yields the same qualitative results as averaging across acceptability derived from each field; including the CA field would simply introduce some additional random noise into the final average acceptability value, obscuring the effects of interest.\footnote{For other studies that use our definition of acceptability, we recommend averaging acceptability across only those fields in which the location of the activation peak is expected to vary systematically across trials. The purpose of this recommendation is to minimize noise in the comparison between simulated and measured acceptability, given that measured acceptability is necessarily influenced by a variety of noise sources which are not explicitly modeled, as mentioned above.}

In our simulation of a canonical alienable possession interpretation of \textit{have} (Figure \ref{fig:activation_history_canonical}), all of the inputs to the conn field are centered at $p = 50$, the center of the field (mid conn). 
Thus, the centroid of the input to the conn field is $\bar{x}_{input} = 50$.
As mentioned above, under this set of inputs, the model will always tend to form an activation peak near the center of the field, leading to high acceptability.
On that particular simulation, the location of the peak was $x_{peak} = 47$, leading to acceptability $= \frac{{1}}{{1 + \lvert 47 - 50 \lvert}} = 0.25$.

\begin{figure}[h]
    \centering
    \includegraphics[width=0.7\linewidth]{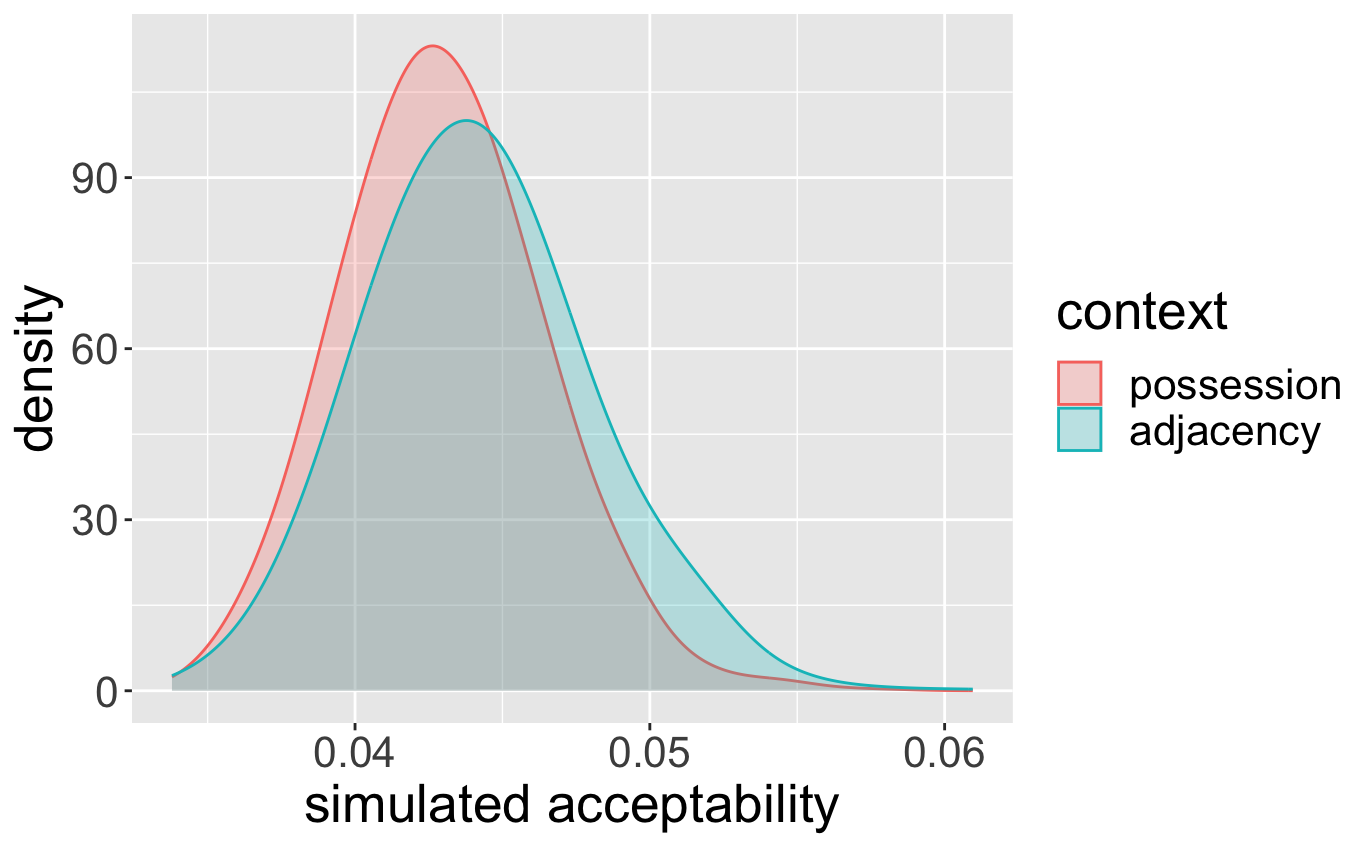}
    \caption{Distribution of simulated acceptability across 1000 simulations in each context.}
    \label{fig:acceptability_distribution}
\end{figure}

Now consider our simulations of a sentence with \textit{have} and two unrelated, inanimate arguments (Figure \ref{fig:activation_history}).
In this case, the centroid of the input to the conn field during the target sentence is slightly to the left of the center of the field ($\bar{x}_{input} = 49.4$), due to the weak low conn input coming from the nouns.
Thus, while both an adjacency interpretation (peak at low conn) and an inalienable possession interpretation (peak at high conn) will lead to relatively low acceptability (since both peaks will be relatively distant from the input centroid), adjacency interpretations will tend to have higher acceptability than inalienable possession interpretations, because the input centroid is slightly to the left of the center, towards the adjacency end of the field.
For example, in the simulations in Figure \ref{fig:activation_history}, acceptability in the adjacency context was 0.047, and acceptability in the possession context was 0.041 (cf. 0.25 for the canonical alienable possession interpretation in Figure \ref{fig:activation_history_canonical}).
These observations---overall low acceptability of \textit{have} sentences with two unrelated, inanimate arguments, but increased acceptability when preceded by an adjacency reading of \textit{have} relative to an inalienable possession reading of \textit{have}---are consistent with measured acceptability ratings \citep{zhang_real-time_2018}.
Figure \ref{fig:acceptability_distribution} displays the distribution of simulated acceptability across the same simulations reported in Figure \ref{fig:peak_distribution}.

\begin{figure}[h]
    \centering
    \includegraphics[width=0.9\linewidth]{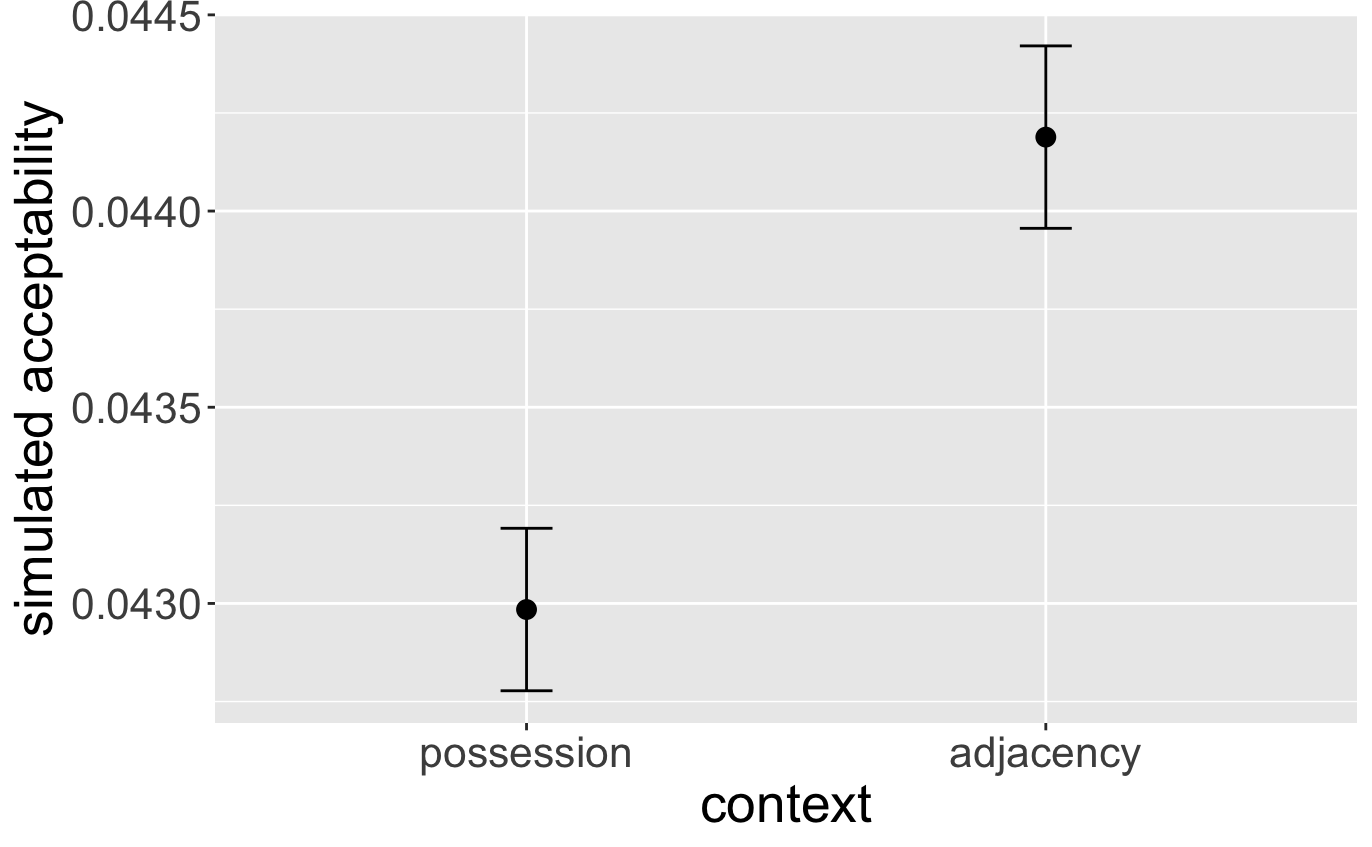}
    \caption{Mean simulated acceptability by context. Error bars indicate 95\% confidence interval.}
    \label{fig:simulated_acceptability}
\end{figure}

It can be seen that there is a high degree of overlap between simulated acceptability ratings in each context, because within each of the regions of the conn field (adjacency or possession), random variability can move the activation peak closer to or further from the input centroid.
However, the distribution of acceptability is shifted higher overall in the adjacency context relative to the possession context, because activation peaks on the left side of the field tend to be closer to the input centroid than peaks on the right side of the field.
To illustrate the by-context difference in acceptability more clearly, Figure \ref{fig:simulated_acceptability} plots mean simulated acceptability by context, which is consistent with the empirical observation \citep{zhang_real-time_2018} that mean acceptability is higher in the adjacency context relative to the possession context.

We note that, while the definition of acceptability in Eq. \ref{eq:acceptability} is a heuristic, rather than a genuine neural dynamic account of acceptability judgments, the definition is generalizable.
Using Eq. \ref{eq:acceptability}, one could calculate acceptability from any set of field inputs (corresponding to the meaning of a sentence) and any (set of) neural activation peak(s) (corresponding to a particular interpretation of a sentence).
Moreover, Eq. \ref{eq:acceptability} may suggest avenues for implementing a neural dynamic model of acceptability judgments, in that the behavior of such a model should exhibit the relation defined here (i.e., higher acceptability corresponds to lower deviation between an activation peak and the input centroid).
Analogous neural mechanisms may be found in the domain of movement error detection.
For example, in the DIVA model of speech production \citep{tourville_diva_2011}, predicted sensory consequences of motor commands \citep[``efference copies":][]{von_holst_reafferenzprinzip_1950} are compared to incoming sensory signals.
A mismatch between predicted and actual sensation leads to detection of an error, and a corresponding increase in neural activation \citep{tourville_neural_2008}.
Intuitions about sentence acceptability may arise from a similar neural mechanism which compares an expected interpretation to an actual interpretation.
Such a mechanism may underlie the electrophysiological difference measured between \textit{have} processing in the two contexts \citep{zhang_real-time_2018}.
We leave it to future work to flesh out the details of this mechanism.

\subsection{Simulated individual variation}
\label{sec:simulated_AQ}
Previous empirical results have also demonstrated that individual variation in the magnitude of the by-context difference in acceptability is predicted by AQ scores, such that individuals with higher AQ scores show a reduced influence of context \citep{zhang_word-meaning_2022}. 
We model individual variation in AQ by varying the parameter $c_{\text{DNF}}$, which controls the magnitude of field-field coupling. 
Stronger field-field coupling is consistent with a greater degree of system-level expectations. 
In other words, given some interpretation on one semantic dimension (e.g., control asymmetry (CA)), individuals can vary in the degree to which they expect a corresponding interpretation on a related semantic dimension (e.g., connectedness (conn)). 
We posit that individuals with higher AQ scores are more influenced by system-level expectations, i.e. stronger field-field coupling. 
With stronger system-level expectations, the processing system is more rigid, and less influenced by real-time information. 
This is consistent with existing findings relating AQ scores to linguistic behavior. 
For example, individuals with higher AQ scores show greater compensation for coarticulation in speech perception \citep{yu_perceptual_2010}. 
In addition, higher AQ individuals are less sensitive to phonetic duration when assigning judgments of prosodic prominence \citep{bishop_individual_2016}.
Both sets of results suggest that higher AQ individuals rely more on system-level expectations (e.g., expected cooccurrence of phonetic signals), and less on the real-time signal itself (e.g., the phonetic duration of a perceived word). 
Stronger excitatory coupling between neural populations is also consistent with the hypothesis that autistic traits correlate with an increased ratio of neural excitation/inhibition \citep{rubenstein_model_2003}.
	
\begin{figure}[h]
    \centering
    \includegraphics[width=0.9\linewidth]{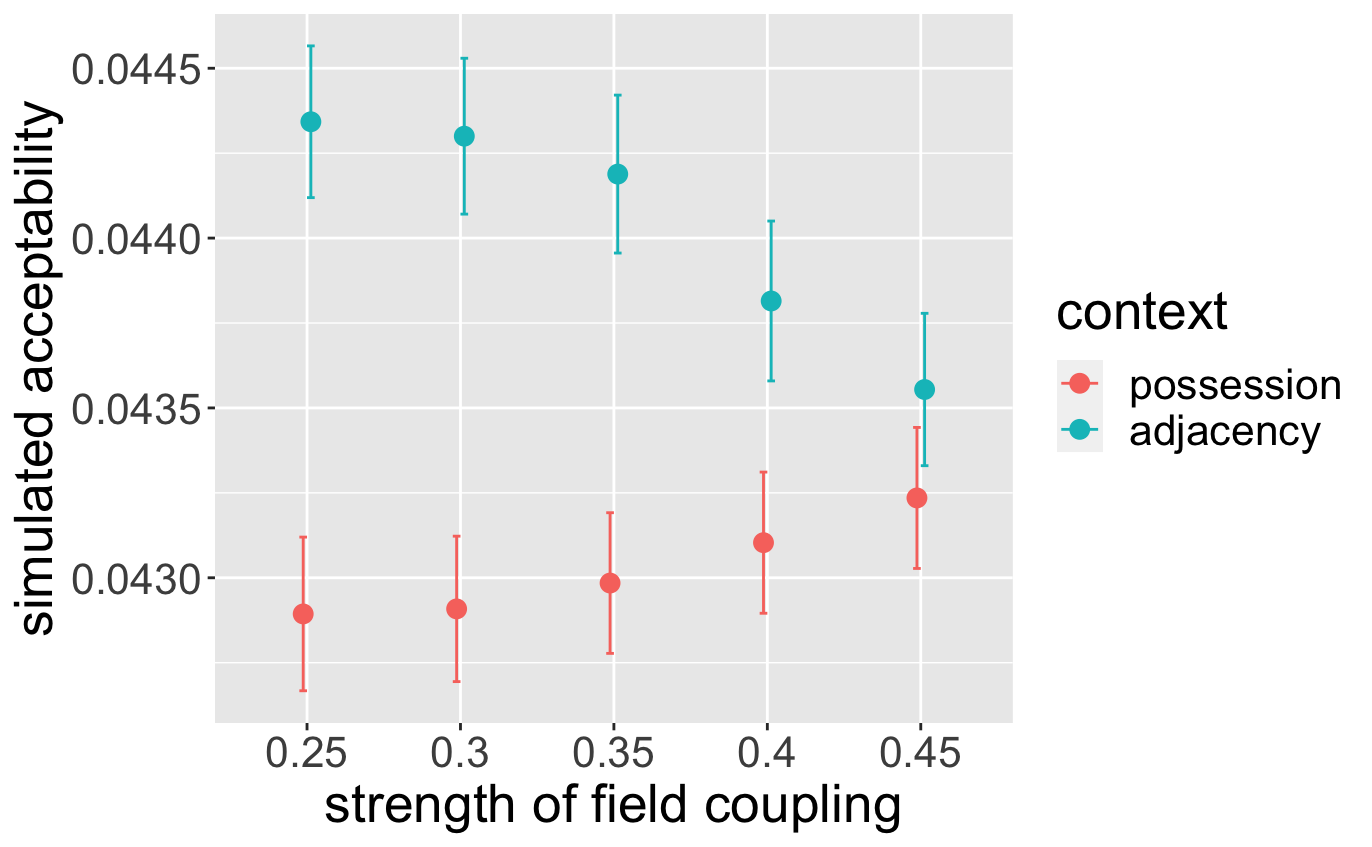}
    \caption{Mean simulated acceptability by context at each level of $c_{\text{DNF}}$. Error bars indicate 95\% confidence interval.}
    \label{fig:vary_coupling_strength}
\end{figure}

In order to apply this hypothesis to the case of contextual influence on the interpretation of \textit{have}, we varied $c_{\text{DNF}}$ from 0.25 to 0.45 in steps of 0.05, and at each level, we ran 1000 simulations in each condition. 
As seen in Figure \ref{fig:vary_coupling_strength}, higher values of $c_{\text{DNF}}$ corresponded with a reduction in the difference in acceptability between conditions. 
In other words, stronger field-field coupling reduced the magnitude of the contextual modulation effect. 
This is because input from the low CA distribution equally favors the low conn and high conn distributions. 
When this input is stronger, the lingering asymmetry between these distributions from \textbf{phase 1} is reduced more quickly, thus reducing the influence of context on interpretation in \textbf{phase 3}.

\subsection{Simulated response time}
\label{sec:simulated_RT}
\begin{figure}[h]
    \centering
    \includegraphics[width=0.9\linewidth]{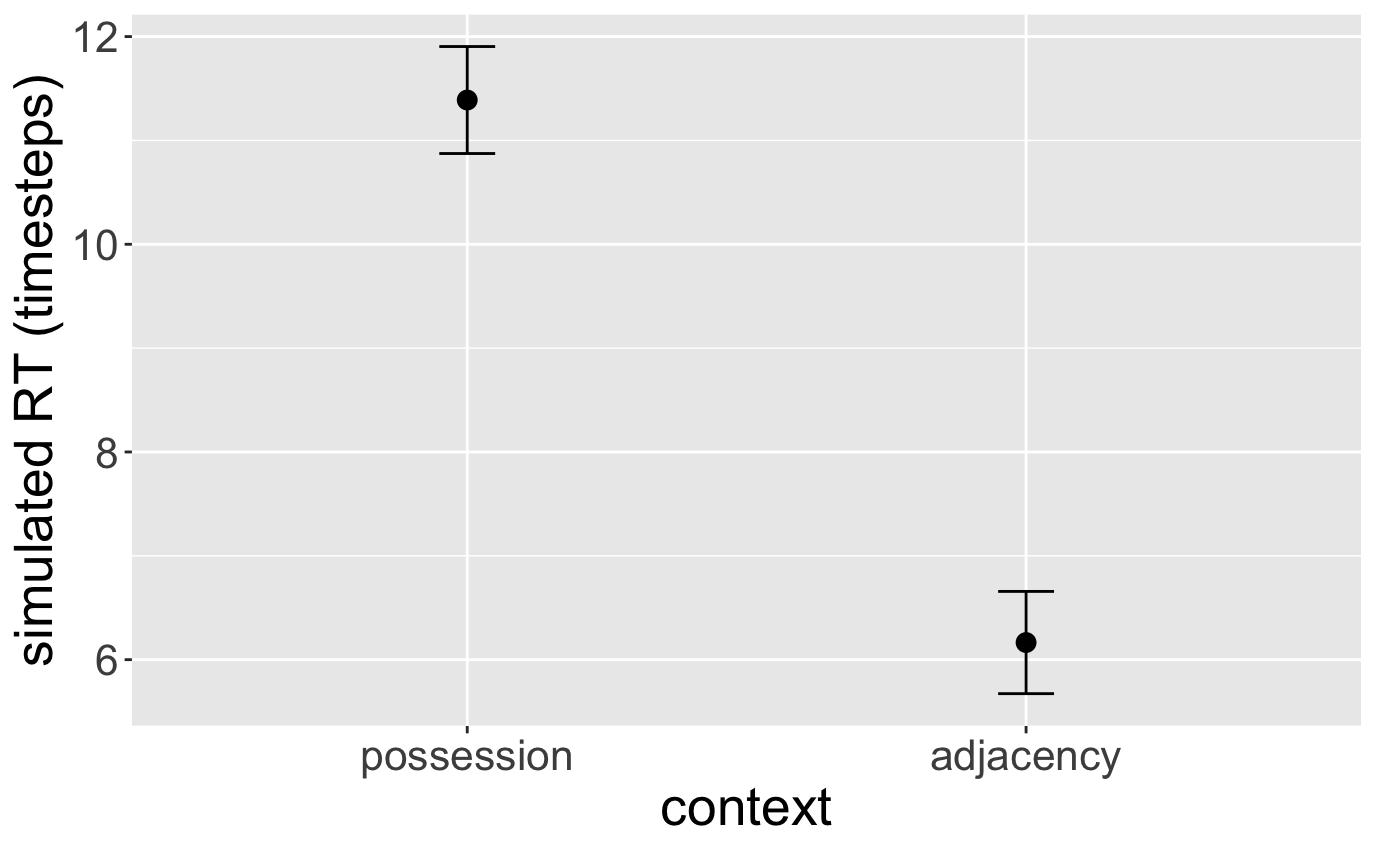}
    \caption{Mean simulated response time by context. Error bars indicate 95\% confidence interval.}
    \label{fig:simulated_RT}
\end{figure}

Next, we examine the influence of context on response time (RT). 
We operationalize RT in each simulation as the number of timesteps between the onset of \textbf{phase 3} and the timestep at which the first neuron passed the threshold for lateral interaction ($u = 0$), leading to activation peak stabilization.
Figure \ref{fig:simulated_RT} displays mean RT by context.

Consistent with self-paced reading data \citep{zhang_real-time_2018}, RT was slower in the possession context relative to the adjacency context.
In the model, the reason for this is as follows.
Across contexts, when an adjacency interpretation occurs (activation peak at low conn), this tends to occur faster (lower RT) than when a possession interpretation occurs (high conn), because of the influence of the weak input to low conn from the nouns.
At the same time, the activation peak is more likely to form at low conn in the adjacency context relative to the possession context, due to the biasing influence of \textbf{phase 1} activation on \textbf{phase 3} activation.
Taken together, this leads to overall slower RT in the possession context relative to the adjacency context.

\begin{figure}[h]
    \centering
    \includegraphics[width=0.9\linewidth]{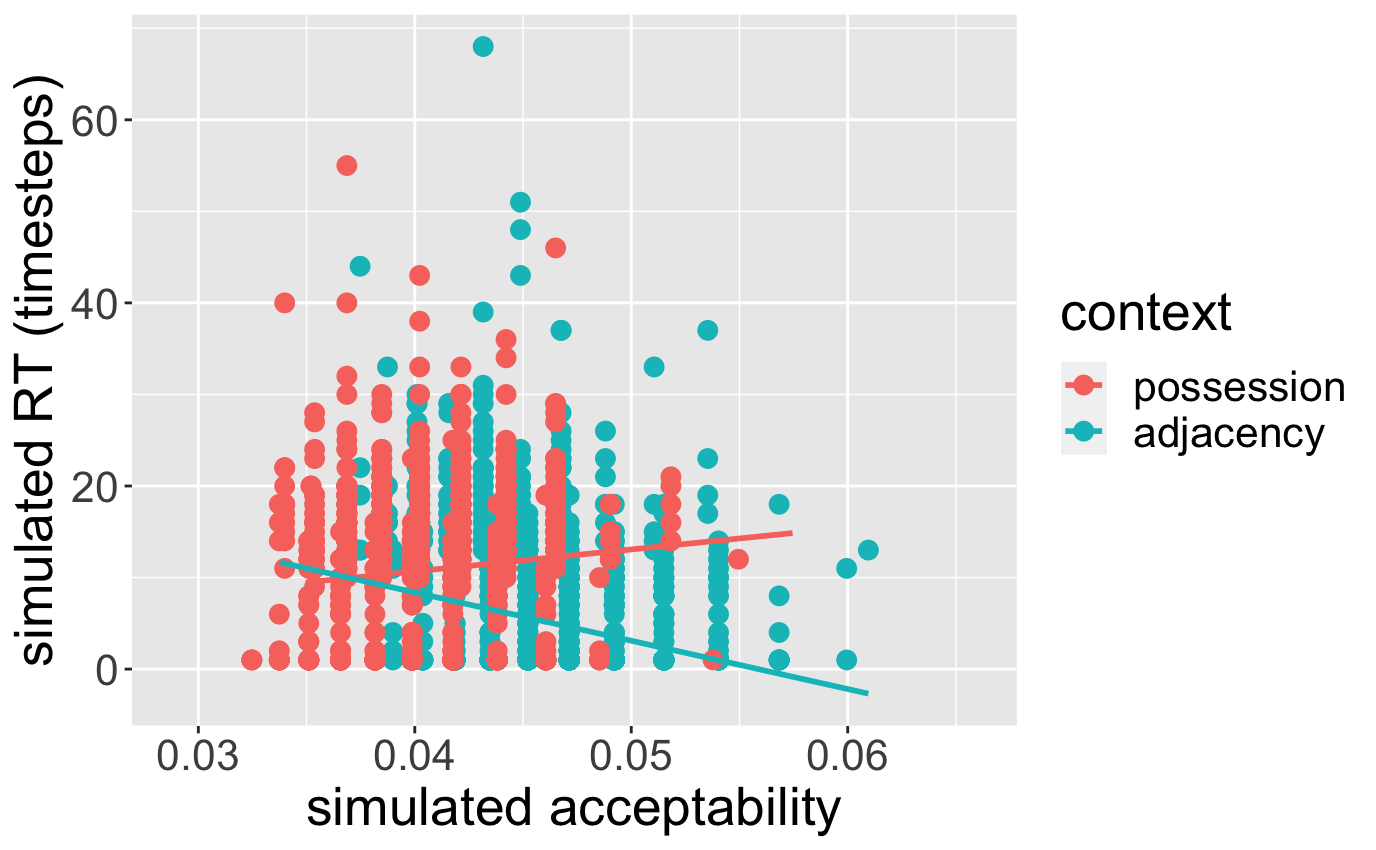}
    \caption{Relationship between acceptability (x-axis) and response time (y-axis) for each simulation in each context (color).}
    \label{fig:RT_by_acc}
\end{figure}

This combination of biases has another interesting consequence: an interaction between context and acceptability in predicting RT, as seen in Figure \ref{fig:RT_by_acc}.
As mentioned above, across contexts, low conn (higher acceptability) interpretations tend to be reached more quickly than high conn (lower acceptability) interpretations, due to the bias from the nouns (weak low conn input).
This leads to an overall negative correlation between acceptability and RT. 
Moreover, in the adjacency context, activation from \textbf{phase 1} (the context sentence) biases activation in \textbf{phase 3} (the target sentence) towards low conn (higher acceptability), so that low conn peaks form even more quickly.
This reinforces the overall negative correlation between acceptability and RT in the adjacency context, as seen in Figure \ref{fig:RT_by_acc} ($\rho = -.26, p < .001$).
In the possession context, on the other hand, activation from \textbf{phase 1} biases activation in \textbf{phase 3} towards high conn (lower acceptability), competing with the overall bias towards low conn from the nouns.
The outcome of this competition is that the correlation between acceptability and RT in the possession context is reversed ($\rho = .16, p < .001$).
Notably, the positive correlation in the possession context is of a weaker magnitude than the negative correlation in the adjacency context, since the correlation in the adjacency context is the result of two reinforcing influences, while the correlation in the possession context is the outcome of two competing influences.

This account is supported by a linear regression model of simulated RT by simulated acceptability (z-scored) and context (sum-coded: possession $= -1$, adjacency $= 1$).
The model revealed main effects of context ($\beta = -2.40$, $SE = 0.18$, $p < .001$) and acceptability ($\beta = -0.51$, $SE = 0.18$, $p < .01$), and an interaction between context and acceptability ($\beta = -1.35$, $SE = 0.18$, $p < .001$).
In order to confirm that the magnitude of the negative correlation in the adjacency context was greater than the magnitude of the positive correlation in the possession context, we multiplied RT values in the possession context by $-1$, and then ran the same linear model as above.
The interaction between context and acceptability remained significant ($\beta = -0.51$, $SE = 0.18$, $p < .01$), confirming a difference in the magnitudes of the correlations, independent of the directions.

\section{Behavioral experiment}
\label{sec:experiment}
To date, studies of contextual facilitation of \textit{have} interpretation have not simultaneously collected data regarding acceptability and processing time. 
In this section, we report an experiment combining acceptability judgments and self-paced reading. 
The purpose of the experiment is twofold: (a) to replicate previous results regarding contextual facilitation of adjacency readings of \textit{have}, as well as individual variation in the magnitude of this effect indexed by the AQ, and (b) to test the model prediction represented in Figure \ref{fig:RT_by_acc}, which has not previously been tested.

\subsection{Experiment design}
\label{sec:design}

\subsubsection{Participants}
\label{sec:participants}

56 adults participated in the experiment (ages 20-30; 32 women, 21 men, 3 nonbinary). 
All participants self-reported that they were native monolingual speakers of American English, and that they had no history of speech, language, hearing, or reading impairment. 
Participants were recruited through Prolific (www.prolific.com). 
Before beginning the experiment, participants provided informed consent under Yale University IRB \#2000033871.

\subsubsection{Materials}
\label{sec:materials}

Each experimental stimulus consisted of a pair of sentences (a context sentence followed by a target sentence) conjoined by \textit{and}.\footnote{All stimuli are included in \ref{appendix:stimuli}.}
Every target sentence was designed to convey an adjacency interpretation of \textit{have}, as in \ref{expLoc}. 

\ex.\label{expLoc} …the oak tree has a skateboard that is red.

It is difficult to construe a possession interpretation of \ref{expLoc} because oak trees do not typically possess skateboards (whether alienably or inalienably). 
Every target sentence had the form ``the [noun1] has a [noun2] that is [adj]". 
[noun2] in the target sentence is the critical word at which an interpretation of \textit{have} can be construed, since after reading [noun2], the participant has read \textit{have} and both of its arguments. 
The relative clause ``that is [adj]" was included as a spillover region. 
There were ten target sentences, each of which was preceded by two different context sentences, for a total of 20 experimental stimuli. 
Each context sentence conveyed either an adjacency reading or an inalienable possession reading of \textit{have}, as in Table \ref{tab:exp_cond}. 
\begin{table}
    \centering
    \begin{tabular}{|c|c|}
    \hline
    \thead{\textbf{adjacency}} & \thead{\textbf{possession}} \\
         \hline
    \makecell{The maple tree has a plastic box \\ behind it and...} & \makecell{The maple tree has a branch \\ that is thick and...} \\
      \hline
    \end{tabular}
    \caption{Example context sentence from each condition, corresponding to target sentence \ref{expLoc}.}
    \label{tab:exp_cond}
\end{table}
Every context sentence had the structure ``the [noun1] has a [noun2] [modifier]". 
[modifier] was either a prepositional phrase (in the adjacency condition) or a ``that is [adj]" phrase (in the possession condition). 
[noun1] was identical between the two conditions in each set, and always contrasted saliently with [noun1] in the target sentence in order to increase overall felicitousness. 
In both conditions, [noun2] in the context was semantically unrelated to [noun2] in the target sentence. 
Moreover, the first phoneme in [noun2] in the context was always different from the first phoneme of [noun2] in the target sentence, in order to minimize confounds from phonological priming. 
All nouns were inanimate in order to maximize the availability of adjacency readings. 
60 filler stimuli were also included. 
20 of the fillers were of the same form as the experimental stimuli, but with contexts that used verbs other than \textit{have} to convey an adjacency reading (10 stimuli) or a possession reading (10 stimuli). 
40 of the fillers were completely unrelated to the experimental stimuli: 20 conveyed interpretations of circumstantial metonymy (e.g., ``the grilled cheese at Table 6 ordered another coffee"), and 20 conveyed non-metonymous counterparts (e.g., ``the customer at Table 6 ordered another coffee"). 
This yielded a total of 80 stimuli (20 experimental + 60 fillers). 
	
In order to examine effects of individual variation in communicative context sensitivity, participants completed the Autism-Spectrum Quotient \citep[AQ:][]{baron-cohen_autism-spectrum_2001}. 
The AQ consists of 50 statements (e.g., ``I prefer to do things the same way over and over again"). 
The participant responds to each statement by selecting one of four options: “definitely disagree”, “slightly disagree”, “slightly agree”, “definitely agree”.

\subsubsection{Procedure}
\label{sec:procedure}

Participants were instructed to complete the experiment in a quiet room where they would be free from distractions for up to an hour. 
To begin each trial, participants clicked on a button at the top of the screen with the words “Click here to begin the next trial”.
Then, the first word of the stimulus appeared in the center of the screen. 
Participants pressed the spacebar to advance to the next word, i.e., word-by-word self-paced reading \citep{just_paradigms_1982}. 
Participants were instructed to read as quickly as possible while making sure to comprehend what they were reading. 
After participants advanced past the last word, they were prompted by the following instruction to give an acceptability rating of the entire stimulus: “How likely would you be to say this sentence, or hear this sentence from another native speaker of English?”. 
Participants gave their response on a seven-point Likert scale (labeled “very unnatural” on the left end and “very natural” on the right) by clicking on the corresponding button at the bottom of the screen. 
	
The experiment began with four practice trials unrelated to the experimental stimuli. 
Then, each stimulus was presented twice in order to increase robustness to noise, given that the data was collected online.\footnote{An anonymous reviewer raises the potential concern of unintended priming at the second presentation of each stimulus. In order to investigate this, we also conducted all analyses separately for each presentation block. The pattern of results reported in Section \ref{sec:results} was mostly equivalent between blocks. In particular, in both blocks, we observed a participant-level correlation between AQ score and context effect on acceptability (see Section \ref{sec:acceptability}), and an interaction between context and acceptability in predicting reading time (see Section \ref{sec:RT_results}). However, there was a difference between blocks in the relationship between context, trial number, and acceptability. The effect of context on acceptability was only significant in block 2 and not block 1, and the effect of trial number on acceptability was only significant in block 1 and not block 2. In order to investigate this pattern further, we ran a single linear mixed effects model with the same structure as the model described in Section \ref{sec:acceptability}, except that it also included a main effect of block, as well as all interactions between block, context, and trial number. Consistent with the model reported in Table \ref{tab:acceptability}, we found main effects of trial number ($\beta = -0.63$, $SE = 0.13$, $p < .001$) and context=adjacency ($\beta = 0.53$, $SE = 0.21$, $p < .05$). We also found a two-way interaction between context=adjacency and trial number ($\beta = 0.41$, $SE = 0.18$, $p < .05$), a two-way interaction between block=2 and trial number ($\beta = 0.58$, $SE = 0.17$, $p < .001$), and a marginal three-way interaction between context=adjacency, block=2, and trial number ($\beta = -0.45$, $SE = 0.23$, $p = .05$). This pattern of results suggests that acceptability ratings in the possession context generally decreased over the course of block 1, before reaching a minimum in block 2. Priming would not explain this effect, since priming should lead to an increase in acceptability in block 2 relative to block 1. Therefore, we interpret the difference between blocks as primarily the result of nonlinearity in the effect of trial number. In the remainder of the paper, we report results from across the two presentation blocks.}
Each of the two blocks was pseudo-randomized such that no two consecutive trials were from the same condition or the same stimulus set. 
The presentation order of the two blocks was counter-balanced between participants. 
Each participant completed 160 trials in total.
After completing the combined self-paced reading and acceptability judgment task, participants completed the AQ. 
The entire procedure was conducted in the same session in Gorilla \citep{anwyl-irvine_gorilla_2020}. 
The session lasted approximately 30-40 minutes.

\subsubsection{Data processing}
\label{sec:processing}

For the analysis of acceptability ratings, trials with acceptability response times greater than 2.5 standard deviations from the participant’s mean were removed. 
This led to the exclusion of 194 trials (2.20\%). 
Before plotting the rating data, raw ratings were z-scored by participant in order to abstract away from idiosyncratic rating styles (e.g., staying towards the ends or towards the center of the scale). 
In statistical models, this was accomplished via random effects by participant. 
For the analysis of word reading times, reading times less than 120 ms or greater than 2000 ms were removed (210 trials, 2.39\%). 
For one participant, 38\% of their trials were removed according to these criteria. 
This participant’s data was subsequently completely excluded from analysis, leaving a total of 2098 experimental trials from 55 participants for analysis. 

Word reading times were log-transformed in order to approximate a normal distribution. 
In order to assess the effects of control variables on reading time, a linear mixed effects model was fit to the log-transformed RT (logRT) of all words with fixed effects of word length (in characters), trial number, and the preceding word’s logRT (all scaled and centered), a random intercept by participant, and random slopes for all predictors by participant. 
The model results are displayed in Table \ref{tab:control_pred}. 
All three control variables were found to significantly affect logRT: words took longer to read when they had more characters or when the preceding word took longer to read.
Words were read faster when the word came later in the experiment.
Thus, rather than analyzing raw logRT, we analyze the residuals of logRT after being regressed, by participant, on the three control predictors described above. 
In other words, we analyze the variance not predicted by the three control predictors. 

\begin{table}
    \centering
    \begin{tabular}{c c c c c c}
    \hline
     & \thead{Estimate} & \thead{Std. Error} & \thead{\textit{df}} & \thead{\textit{t} value} & \thead{\textit{p} value} \\
         \hline
         \thead{(Intercept)} & $5.60$ & $0.03$ & $53.87$ & $198.29$ & $< .001$ \\
         \thead{char\_length} & $0.01$ & $0.00$ & $54.07$ & $4.23$ & $< .001$ \\
         \thead{trial\_num} & $-0.11$ & $0.01$ & $53.93$ & $-15.66$ & $< .001$ \\
         \thead{prec\_logRT} & $0.13$ & $0.01$ & $53.92$ & $19.66$ & $< .001$ \\
         \hline
    \end{tabular}
    \caption{Effects of control predictors on logRT.}
    \label{tab:control_pred}
\end{table}

Each AQ response was initially coded on a four-step Likert scale where ``definitely disagree” = 1 and ``definitely agree” = 4. 
For half of the questions on the AQ, agreement signifies greater autistic traits and disagreement signifies reduced autistic traits; this is reversed for the other half of questions. 
For questions where disagreement signified greater autistic traits, we reversed the numerical response so that higher values always corresponded to greater autistic traits. 
Thus, individual AQ scores could range from 50 (very low autistic traits) to 200 (very high autistic traits).

\subsection{Experiment results}
\label{sec:results}

\subsubsection{Acceptability}
\label{sec:acceptability}

In order to assess the effect of context on acceptability, we fit nested linear mixed effects models to acceptability ratings using the \textit{lme4} package \citep{bates_fitting_2015} in R \citep{r_core_team_r_2021}. 
All models included random intercepts by item and by participant, and random slopes for context by item and by participant. 
Model comparison revealed that a control fixed factor for trial number (scaled and centered) significantly improved model fit over a baseline model that only included random effects ($\chi^2(1) = 14.67$, $p < .001$).
Additionally including the experimental fixed factor of context (treatment coded; reference level = possession) significantly improved model fit over the control model ($\chi^2(1) = 5.27$, $p < .05$). 
The results of the full model are displayed in Table \ref{tab:acceptability}. 
Trial number significantly decreased acceptability ratings, such that ratings generally decreased over the course of the experiment. 
Regarding the experimental factor of interest, ratings were significantly higher in the adjacency context relative to the possession context.

\begin{table}
    \centering
    \begin{tabular}{c c c c c c}
    \hline
     & \thead{Estimate} & \thead{Std. Error} & \thead{\textit{df}} & \thead{\textit{t} value} & \thead{\textit{p} value} \\
         \hline
         \thead{(Intercept)} & $3.23$ & $0.25$ & $28.77$ & $12.96$ & $< .001$ \\
         \thead{trial number} & $-0.11$ & $0.03$ & $1628.44$ & $-3.88$ & $< 0.001$ \\
         \thead{context=adjacency} & $0.28$ & $0.12$ & $29.07$ & $2.36$ & $< .05$  \\
         \hline
    \end{tabular}
    \caption{Linear mixed effects model of acceptability ratings.}
    \label{tab:acceptability}
\end{table}

\begin{figure}[h]
    \centering
    \includegraphics[width=.9\linewidth]{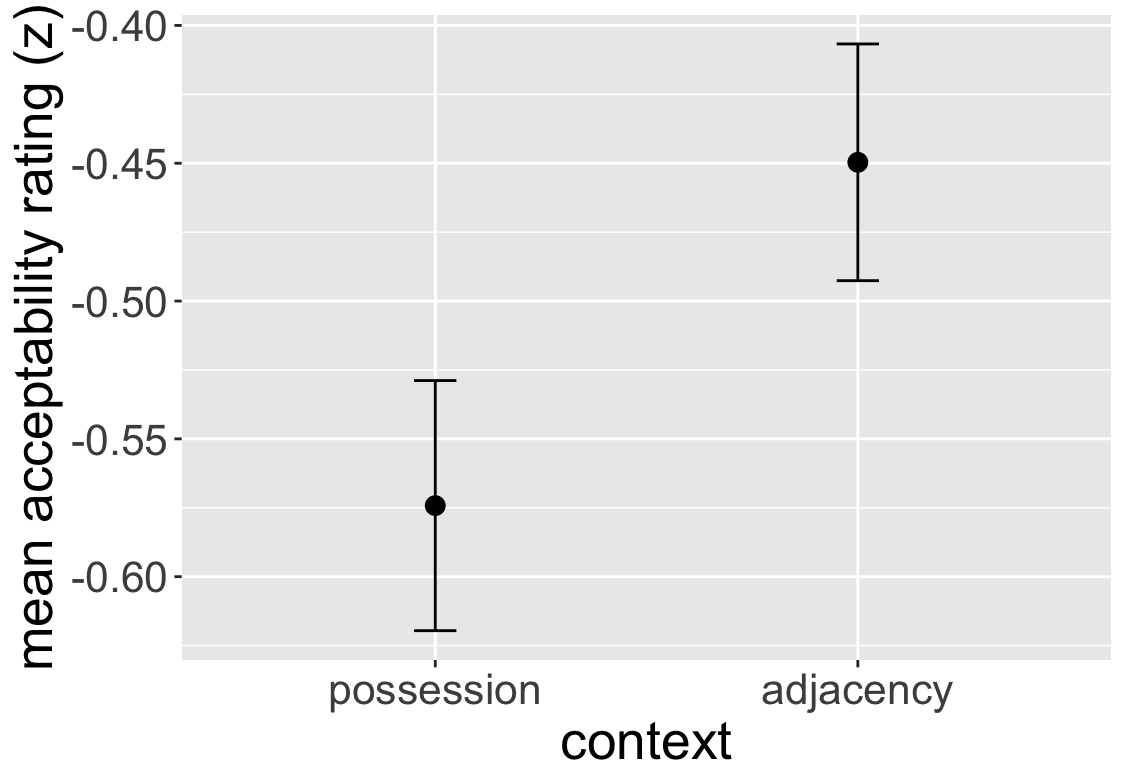}
    \caption{Mean z-scored (by participant) acceptability ratings by context. Error bars indicate 95\% confidence interval. Compare to simulated results in Figure \ref{fig:simulated_acceptability}.}
    \label{fig:acceptability}
\end{figure}

Figure \ref{fig:acceptability} displays mean z-scored (by-participant) acceptability ratings in each context. 
Both means are below 0, suggesting that the experimental stimuli were generally less acceptable than the filler stimuli, consistent with the general markedness of the target sentences \citep{zhang_real-time_2018}, which was also evident in simulated acceptability (Section \ref{sec:simulated_acceptability}). 
Consistent with the results of the regression model, ratings were higher in the adjacency context relative to the possession context. 

\begin{figure}[h]
    \centering
    \includegraphics[width=.9\linewidth]{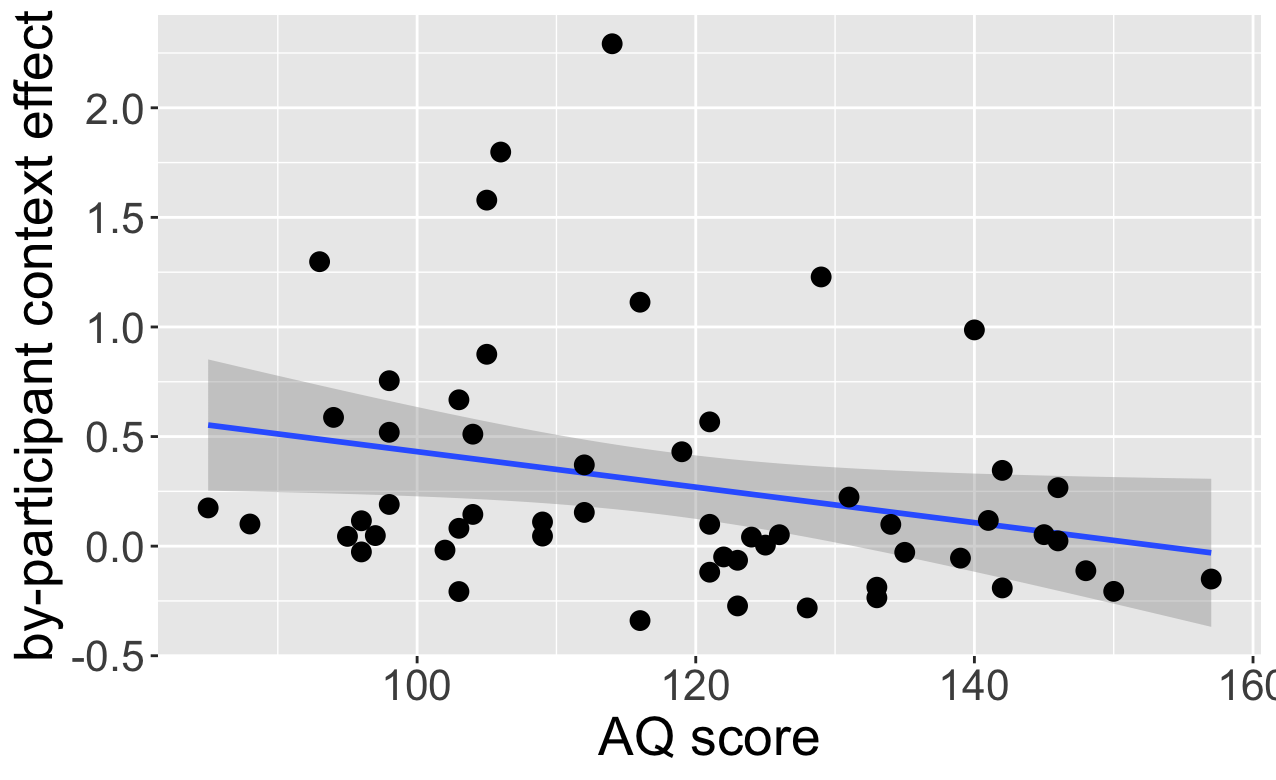}
    \caption{Relationship between by-participant context effect and AQ score.}
    \label{fig:rating_by_AQ}
\end{figure}

In order to assess whether individual variation in the magnitude of contextual facilitation is predicted by AQ scores, we plot the by-participant slopes for the effect of context against AQ score in Figure \ref{fig:rating_by_AQ}. 
A Spearman test confirms a negative correlation ($\rho = -.36$, $p < .01$): subjects with higher AQ scores showed a smaller effect of context, i.e. reduced contextual facilitation of adjacency \textit{have} sentences.

\begin{figure}[h]
    \centering
    \includegraphics[width=.9\linewidth]{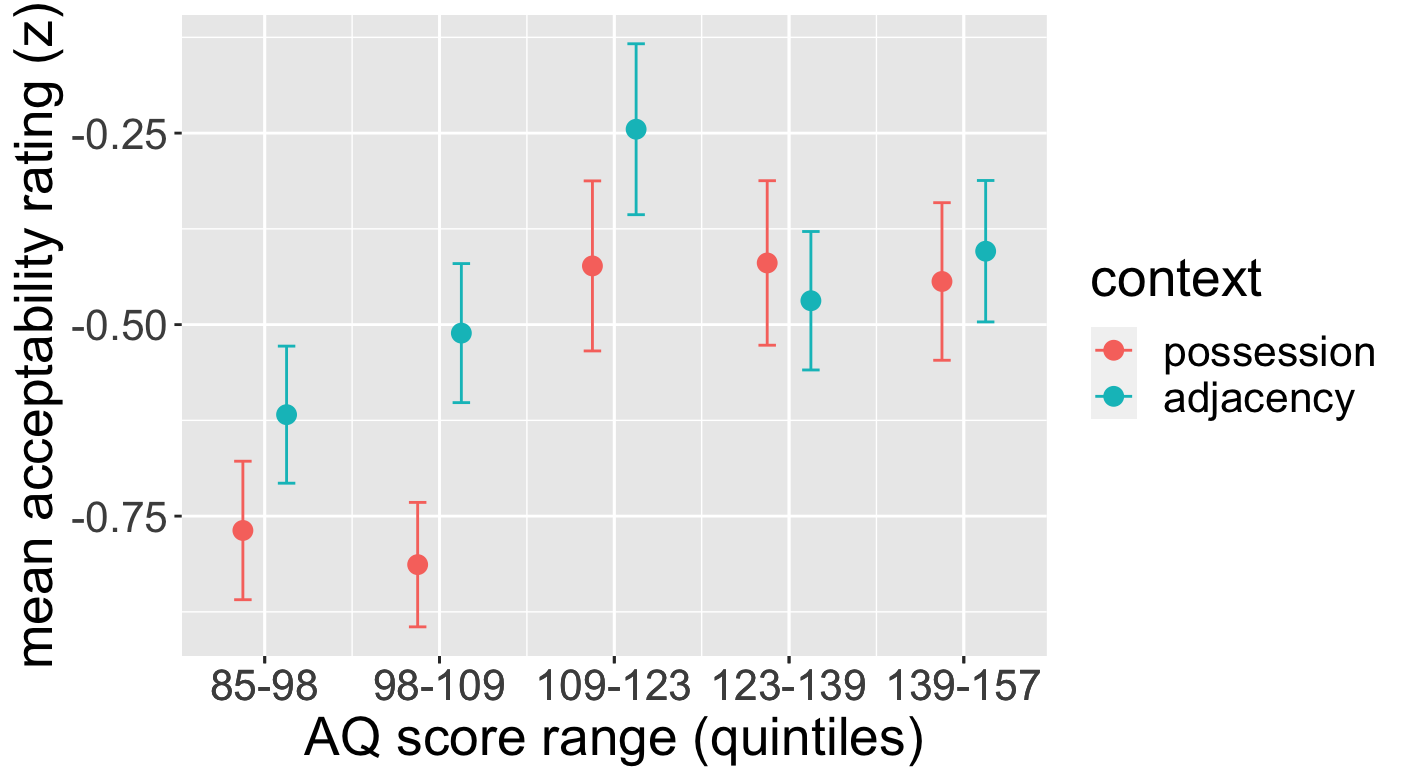}
    \caption{Relationship between condition means and participant AQ scores, binned into quintiles. Error bars indicate 95\% confidence interval. Compare to simulated results in Figure \ref{fig:vary_coupling_strength}.}
    \label{fig:rating_by_AQ_quintiles}
\end{figure}

This trend can also be seen in Figure \ref{fig:rating_by_AQ_quintiles}, which plots mean rating (z-scored by participant) by condition against participant AQ scores, binned into quintiles, analogous to Figure \ref{fig:vary_coupling_strength} which displays the simulation results.
The difference in mean rating between conditions tends to decrease as AQ score increases, although this trend is not as clean as in the simulation results.
We revisit this result in Section \ref{sec:experiment_discussion}.

\subsubsection{Reading time}
\label{sec:RT_results}
Next, we turn to the reading time results.
We fit nested linear mixed effects models to the summed residualized log-transformed reading times (see Section \ref{sec:processing}) of the critical word [noun2] and each of the words in the spillover region ``that is [adj]".
All models included random intercepts by item and by participant.\footnote{Random slopes were not included because they led to model convergence issues and were also not of theoretical interest in this case.}
Using the same methods for fitting nested linear mixed effects models as above (Section \ref{sec:acceptability}), we assessed the influence of the following fixed factors on reading time: context (sum-coded: possession $= -1$, adjacency $= 1$), acceptability (z-scored by subject), and their interaction.
Adding a fixed factor of context did not lead to a significant improvement in fit over a baseline model with only random effects ($\chi^2(1) = 0.33$, $p = .57$).
However, adding a fixed factor for acceptability rating did lead to a significant improvement in fit over the baseline model ($\chi^2(1) = 5.64$, $p < .05$).
Adding a fixed factor for context back in did not lead to an improvement in fit over the model with only a fixed factor for acceptability ($\chi^2(1) = 0.64$, $p = .42$).
Importantly, however, adding an interaction term did lead to a significant improvement in fit over the model with only fixed factors for main effects ($\chi^2(1) = 4.58$, $p < .05$).
We report the results of the full model with both main effects and their interaction in Table \ref{tab:RT_by_acc}.

\begin{table}
    \centering
    \begin{tabular}{c c c c c c}
    \hline
     & \thead{Estimate} & \thead{Std. Error} & \thead{\textit{df}} & \thead{\textit{t} value} & \thead{\textit{p} value} \\
         \hline
         \thead{(Intercept)} & $-0.02$ & $0.03$ & $30.86$ & $-0.75$ & $.46$ \\
         \thead{context} & $-0.01$ & $0.02$ & $2061$ & $-0.59$ & $.55$ \\
         \thead{acceptability} & $-0.05$ & $0.02$ & $1554$ & $-2.51$ & $< .05$  \\
         \thead{context:acceptability} & $-0.04$ & $0.02$ & $2089$ & $-2.14$ & $< .05$  \\
         \hline
    \end{tabular}
    \caption{Linear mixed effects model of residualized reading time for the critical region.}
    \label{tab:RT_by_acc}
\end{table}

\begin{figure}[h]
    \centering
    \includegraphics[width=1\linewidth]{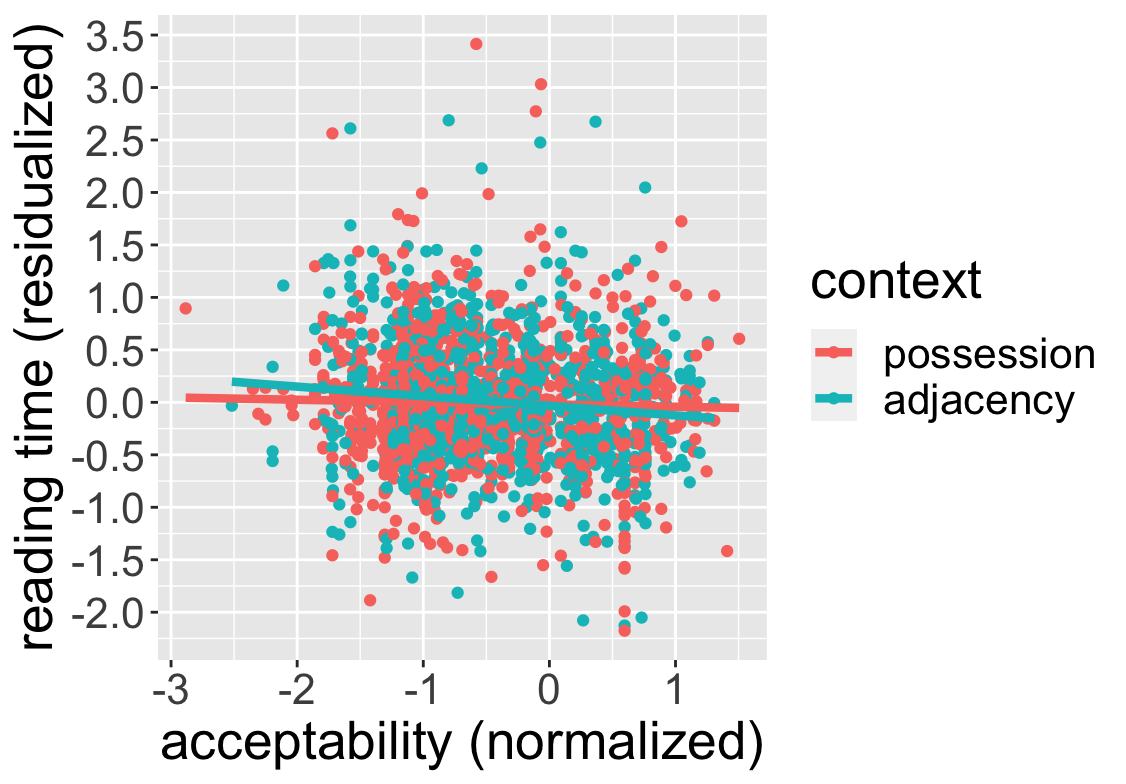}
    \caption{Mean residualized log-transformed reading times of the critical region (y-axis) by normalized acceptability rating (x-axis) and context (color).}
    \label{fig:RT_by_acc_exp}
\end{figure}

Consistent with the model simulation results described in Section \ref{sec:simulated_RT}, we observed a main negative effect of acceptability on reading time.
Also consistent with the simulation results, we observed an interaction between context and acceptability in predicting reading time: the relationship between acceptability and reading time was more negative in the adjacency context relative to the possession context (Figure \ref{fig:RT_by_acc_exp}).
Finally, as predicted by the model, we observed a significant negative correlation between acceptability and reading time in the adjacency context ($\rho = -.10, p < .01$).

\begin{figure}[h]
    \centering
    \includegraphics[width=1\linewidth]{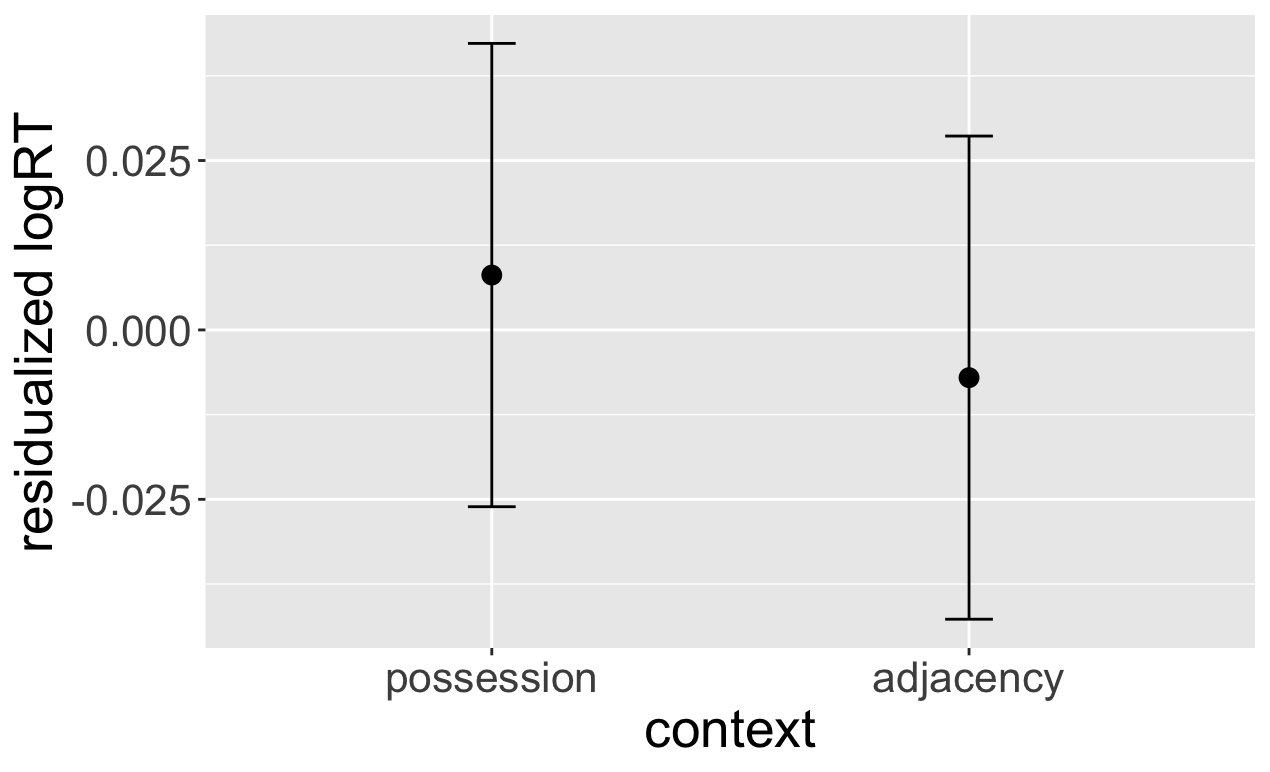}
    \caption{Mean residualized log-transformed reading times of the critical region by context. Error bars indicate 95\% confidence interval. Compare to simulated results in Figure \ref{fig:simulated_RT}.}
    \label{fig:RT_region}
\end{figure}

\begin{figure}[h]
    \centering
    \includegraphics[width=1\linewidth]{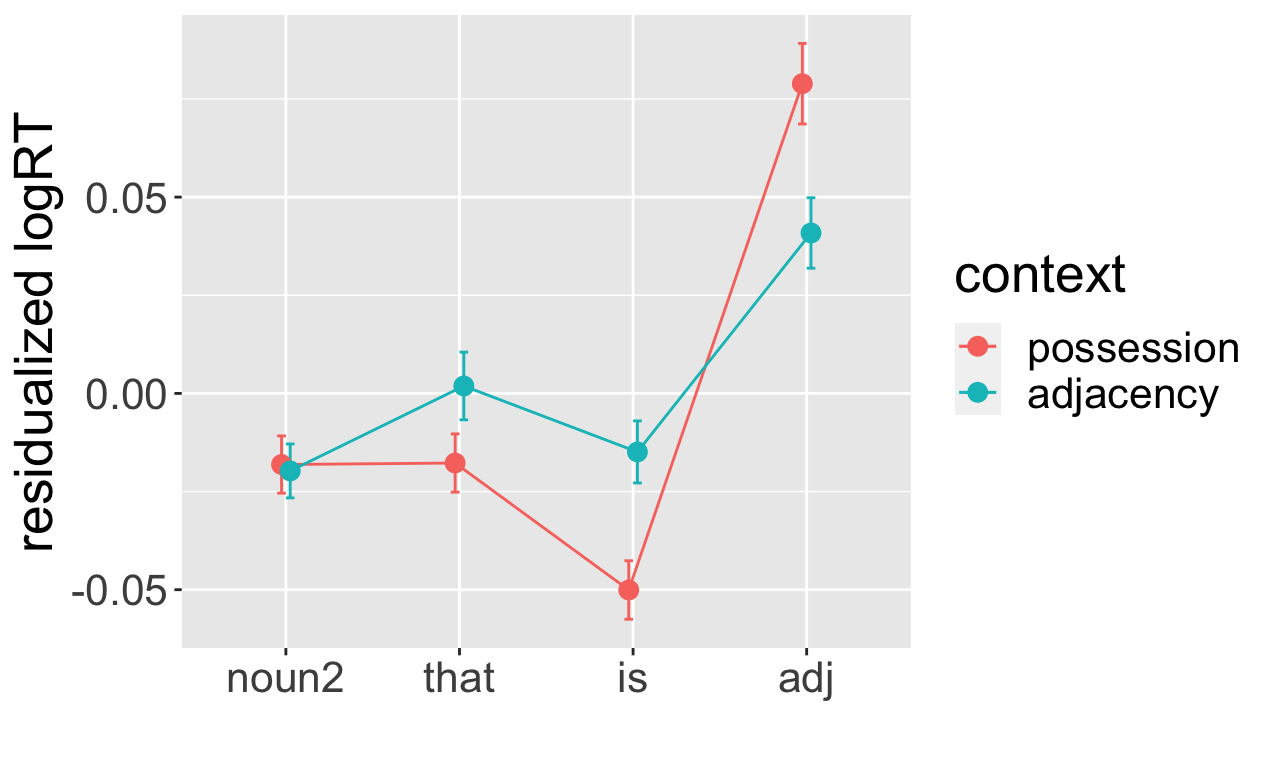}
    \caption{Mean residualized log-transformed reading times of each word in the critical region by context. Error bars indicate 95\% confidence interval.}
    \label{fig:RT_word}
\end{figure}

Two effects predicted by the model were not observed.
First, we did not observe the predicted positive correlation between acceptability and reading time in the possession context ($\rho = -.01, p = .76$).
We discuss a possible explanation of this discrepancy between prediction and observation in Section \ref{sec:summary_discussion}. 
Second, we did not observe a main effect of context on reading time (Figure \ref{fig:RT_region}).
This result contrasts with the prediction from the simulations, as well as previous results \citep{zhang_real-time_2018}.
In order to investigate this discrepancy, we examined the reading time of each word in the critical region, displayed in Figure \ref{fig:RT_word}.
We fit separate linear mixed effects models to residualized log-transformed reading time for each word, with the same structure as the model reported in Table \ref{tab:RT_by_acc}.
There was no main effect of context on RT for the critical word [noun2] ($\beta = 0.00$, $SE = 0.01$, $p =.70$) or the following word ``that" ($\beta = 0.00$, $SE = 0.01$, $p =.51$).
However, at the verb ``is" (two words after the critical word), there was a main effect of context on RT in the opposite direction to the prediction ($\beta = 0.02$, $SE = 0.01$, $p <.01$).
Not until [adj], three words after the critical word, did we observe a main effect of context on RT in the predicted direction ($\beta = -0.03$, $SE = 0.01$, $p <.001$).
We discuss a possible explanation for this pattern in Section \ref{sec:experiment_discussion}.

\subsection{Discussion}
\label{sec:experiment_discussion}
The acceptability results replicate previously reported results regarding contextual facilitation of adjacency \textit{have} readings (Figure \ref{fig:acceptability}), as well as individual variation in the magnitude of facilitation (Figures \ref{fig:rating_by_AQ}, \ref{fig:rating_by_AQ_quintiles}). 
The qualitative similarity between the effect of AQ score on measured acceptability ratings (Figures \ref{fig:rating_by_AQ}, \ref{fig:rating_by_AQ_quintiles}) and the effect of field coupling strength on simulated acceptability (Figure \ref{fig:vary_coupling_strength}) provides some support for our proposal that measured AQ scores are related to field coupling strength (see Section \ref{sec:simulated_AQ}).

With respect to reading time, the predicted main effect of context on reading time was observed on the last word in the critical region; at the previous word (``is"), the \textit{opposite} effect was observed (Figure \ref{fig:RT_word}).
We hypothesize that the unpredicted effect on ``is" was due to unintended lexical priming in the possession context, since both the possession context sentence and the target sentence included a relative clause beginning with ``that is”, while the adjacency context sentence did not include this phrase. 
This low-level similarity between the possession context sentence and the target sentence, which was not present in the stimuli of \cite{zhang_real-time_2018}, likely obscured the main effect of context on RT when measured across the entire critical region. 
This said, our replication of the acceptability effect and the by-participant correlation between the acceptability effect and AQ score, as well as our confirmation of the predicted interaction between context and acceptability in predicting reading time, suggests that this difference in stimulus design did not otherwise confound the results of the experiment. 

Finally, the predicted relationship between context, acceptability, and reading time was partially supported (Table \ref{tab:RT_by_acc}, Figure \ref{fig:RT_by_acc_exp}). 
As predicted, there was a significant main effect of acceptability on reading time, a significant interaction between context and acceptability in predicting reading time, and a significant negative correlation between reading time and acceptability in the adjacency context.
However, the predicted positive correlation between reading time and acceptability in the possession context was not observed.
We discuss a possible explanation for this unpredicted result in Section \ref{sec:summary_discussion}.

\section{General discussion \& conclusion}
\label{sec:general_discussion}

\subsection{Summary \& discussion}
\label{sec:summary_discussion}

We have argued for a dynamic neural model of lexical meaning and demonstrated its behavior using the English lexical item \textit{have} as a test case. 
The cognitive basis of the neural model is a continuous meaning space with two parameters: control asymmetry and connectedness. 
Interpretations of ``possession" and ``adjacency" associated with \textit{have} result from the neural dynamics governing interpretation along those dimensions. 
The apparent discreteness of the different readings results from the structure imposed by node-field coupling and field-field coupling, along with the property of nonlinearity in the neural dynamics.

In the model, lexical meaning is a coupling pattern between a neural node representing the lexical item and dynamic neural fields (DNFs) governing interpretation on continuous semantic dimensions. 
Interpretation occurs in time as activation of the lexical node causes peaks of activation in the semantic DNFs.
The locations of the peaks in feature space correspond to the content of the interpretation.
Dependencies between semantic dimensions, schematized in Figures \ref{fig:MDS_sparse} and \ref{fig:MDS}, are modeled as coupling patterns between DNFs, shown in Figure \ref{fig:DNF_coupling}.

Simulations from the model captured known empirical effects.
In particular, the specific reading evoked by English \textit{have} was influenced by preceding context.
While adjacency readings were overall more likely than possession readings (because the two nouns in the target sentence were unrelated and both inanimate), the likelihood of a possession reading was increased following a possession context, analogous to syntactic priming effects in comprehension \citep[e.g.,][]{arai_priming_2007,ledoux_syntactic_2007,tooley_syntactic_2010}.
Moreover, adjacency readings were generally reached more quickly than possession readings, as reflected by the main effect of context on response time.
We also related the simulation results to acceptability judgments through a generalizable measure (Eq. \ref{eq:acceptability}), deriving the empirical effect of context on acceptability \citep{zhang_real-time_2018}.
The effects of the context sentence on target sentence interpretation occurred in the model because activation states persist in time, continuing to bias interpretation until they return to a resting state.\footnote{The model parameter $\tau$, which controls the rate of field evolution, modulates the temporal extent of contextual bias on interpretation.}

Model simulations also exhibited covariation between the magnitude of contextual modulation and the strength of coupling between DNFs.
Stronger coupling makes DNFs more resistant to the effects of context.
In other words, there is a stronger influence of long-term knowledge, relative to immediate context, on lexical interpretation.
Motivated by previously reported observations regarding individual variation in speech behavior (described in Section \ref{sec:simulations}), we related DNF coupling strength to the Autism-Spectrum Quotient (AQ).
Our proposal that individual variation in the strength of coupling between DNFs is indexed by the AQ generates testable predictions, including in domains unrelated to linguistic meaning.
For example, in an experimental task that requires learning associations between object color (one DNF) and object shape (another DNF), individuals with higher AQ scores are predicted to show greater surprisal effects (e.g., slowed response times) when encountering an object that violates the learned associations. 
Modeling learned dependencies between dimensions as DNF coupling may also shed light on other issues in linguistics.
For example, phonological inventories (possible sounds in a language) and phonotactic constraints (possible sound sequences in a language) may be explained as the result of coupling between DNFs representing phonetic dimensions.
Under this view, DNF coupling patterns are an important component of language-specific knowledge, and thus an important dimension of variation between speakers of different languages. 
Among speakers of the same language, processes conditioned by phonological knowledge, such as accent and perceptual illusions \citep{davidson_sources_2012,dupoux_epenthetic_1999,halle_dental--velar_2007,kabak_perceptual_2007}, are predicted to covary with AQ scores.

Finally, under our proposed definition of acceptability (see Section \ref{sec:simulated_acceptability}), model simulations predicted an interaction between context and acceptability in predicting response time (see Figure \ref{fig:RT_by_acc}). 
This prediction has not previously been reported or empirically tested (although other models may generate the same prediction under certain assumptions; see Section \ref{sec:bayesian_models}).
Our model generated this prediction in the following way.
Recall that, during target sentence processing, activation trajectories in the conn DNF were generally more likely to converge to an activation peak at the low end of the DNF (i.e., an adjacency interpretation), due to the weak input from the two nouns in the target sentence (see Section \ref{sec:simulations}).
In other words, independent of the influence of the context sentence, the adjacency attractor was slightly stronger than the possession attractor.
For this reason, there was a main negative effect of simulated acceptability on simulated response time, such that when a more acceptable (adjacency) reading was reached, it tended to be reached more quickly.
The context sentence additionally biased target sentence processing by exerting an influence on the initial neural activation state at the onset of target sentence processing.
The adjacency context sentence reduced the distance in state space between the initial activation state and the adjacency attractor.
This reinforced the overall bias for activation trajectories to converge to the adjacency attractor, magnifying the overall negative relationship between acceptability and response time.
At the end of the possession context sentence, on the other hand, the neural activation state in the conn DNF was closer in state space to the possession attractor.
This increased the likelihood of convergence to the possession attractor, competing with the overall bias towards the adjacency attractor.
As a result, in the possession context, there was a \textit{positive} correlation between simulated acceptability and simulated response time, such that when a more acceptable (adjacency) reading was reached, it tended to be reached more \textit{slowly}.
However, the magnitude of the positive correlation in the possession context was of a weaker magnitude than the negative correlation observed in the adjacency context. 
The reason for this is that the correlation in the adjacency context reflects the combination of two reinforcing influences, while the correlation in the possession context reflects a conflict between two competing influences.

The predicted relationship between context, acceptability, and response time was partially confirmed by the results of a behavioral experiment combining self-paced reading and acceptability judgments. 
Specifically, results showed an interaction between context and acceptability in predicting reading time on a trial by trial basis, such that the relationship between acceptability and reading time was more negative in the adjacency context relative to the possession context (Figure \ref{fig:RT_by_acc_exp}).
Also as predicted, there was a significant negative main effect of acceptability on reading time, and a negative correlation between reading time and acceptability in the adjacency context.

However, the predicted positive correlation between reading time and acceptability in the possession context was not observed.
We attribute this discrepancy to a bias introduced by the experimental task.
While our simulated acceptability measure is derived only from the interpretation of the target sentence (which is still, of course, influenced by the context sentence), participants were instructed to rate the acceptability of both sentences together.
The perceived (in)congruity of the context and target sentences may have increased the magnitude of the negative correlation between acceptability and reading time across both contexts, eliminating the predicted (but already weak) positive correlation in the possession context, since this potential factor was not incorporated in our model simulations.
This said, the key finding, predicted by the model, is that context modulates the relationship between acceptability and reading time. 
We leave for future work the more detailed neural modeling of acceptability judgments (see Section \ref{sec:simulated_acceptability}) that may help shed light on this issue.

\subsection{Broader theoretical context}
\label{sec:broader_theoretical_context}
\subsubsection{Dynamical systems models of sentence processing}
\label{sec:dynamical_systems_models}
Our model is consistent with a tradition of applying dynamical systems theory to human language, including the DFT work cited in Section \ref{sec:DFT}, but also work that predates DFT or does not rely on the specific neural mechanisms of DFT. 
\cite{poston_mister_1987} and \cite{wildgen_ambiguity_1995}, for example, proposed that lexical semantic ambiguities are the result of multistable energy landscapes in the dynamics of cognitive states defined by continuous semantic dimensions.
While remaining agnostic about the neural mechanisms that give rise to the multistabilities, they rely on the same concepts that we rely on here: continuous semantic dimensions, apparent categoricity through nonlinear dynamics, and influence of context via persistence through time of cognitive states.
Our model can be seen as a more detailed extension of this work, incorporating the specific neural mechanisms from DFT and thus deriving a range of specific quantitative predictions.

Related work has applied dynamical systems ideas to language processing through the lens of connectionist or parallel distributed processing models \citep[e.g.,][]{kawamoto_nonlinear_1993,tabor_parsing_1997,tabor_dynamical_1999}.
The input units of these models are discrete localist representations, in contrast to the continuous representations in our model.
In \cite{kawamoto_nonlinear_1993}, input units represent letters, phonemes, parts of speech, and dummy semantic variables which capture semantic similarities between lexical items.
In \cite{tabor_parsing_1997}, input units represent words.
However, after training, network parameters form a high-dimensional continuous similarity space, between lexical items in \cite{kawamoto_nonlinear_1993}, and between grammatical classes in \cite{tabor_parsing_1997}.

These models are similar to ours in a variety of ways: sentence processing corresponds to a trajectory through a continuous state space; linguistic interpretations correspond to attractors in the state space; and response time corresponds to the temporal duration from onset of stimulus to settling at an attractor. 
However, there are also crucial differences.
First, the state spaces in \cite{kawamoto_nonlinear_1993} and \cite{tabor_parsing_1997} are only continuous in \textit{activation} (and related variables like connection weights and biases). 
The units over which these continuous variables are defined are themselves discrete, as described above. 
In our model, both activation and the units of representation over which activation is defined are continuous. 
Relatedly, the dimensions of the state space in our model are interpretable, and much fewer than those of \cite{kawamoto_nonlinear_1993} and \cite{tabor_parsing_1997}.

Moreover, the processing dynamics of our model are independently motivated.
This requires less stipulation in order to link model behavior to human behavioral data.
By contrast, \cite{tabor_parsing_1997}, for example, specifies a mechanism of gravitational attraction to locations in state space in order to derive response time predictions.
This mechanism is separate from the processing network itself, and is not related to any specific neural mechanism or principle.
In our model, like that of \cite{kawamoto_nonlinear_1993}, attractors in state space arise from the dynamics of the processing model itself, and so no separate mechanism has to be stipulated in order to link model behavior to human response times.
Our model represents a further improvement over \cite{kawamoto_nonlinear_1993} in that connections between units are constrained by neural principles (in particular, local excitation and distal inhibition), rather than being free to vary during training.

An advantage of connectionist models is that the structure of the state space (representing lexical items in \cite{kawamoto_nonlinear_1993} and syntactic classes in \cite{tabor_parsing_1997}) emerges through learning. 
We do not explicitly model the emergence of the state space structure; rather, we stipulate the structure (node-field coupling and field-field coupling) based on prior empirical evidence. 
This does not represent a commitment on our part that this structure does not ultimately emerge through learning; differences between languages necessitate the conclusion that some coupling relationships are formed during learning. 
However, it is possible that other coupling relations are innate, as suggested by the cross-linguistic similarities described in Section \ref{sec:continuous_space}. 
By stipulating the parameters of our model, we sidestep the role of learning in this paper, focusing instead on processing in the adult system.
This said, the tools exist in the DFT framework for modeling parameter learning over developmental timescales \citep[e.g.,][]{bhat_word-object_2022}. 
Applying these tools to the learning of lexical polysemy would be a fruitful area for future work. 

\subsubsection{Bayesian models of sentence processing}
\label{sec:bayesian_models}
Models of sentence comprehension based on Bayesian inference, like surprisal theory \citep{levy_expectation-based_2008}, utilize probability distributions over possible representations; the distributions update in response to linguistic input. 
By stipulating that the time to comprehend a word is proportional to the Kullback-Leibler divergence between the probability distributions before and after encountering the word, surprisal theory is able to predict a variety of subtle empirical patterns in reading times. 
In \cite{levy_expectation-based_2008}, probability distributions are defined over syntactic structures, which are inherently discrete. 
Yet, it should be straightforward to extend this framework to lexical semantics: probability distributions would be defined over a continuous space of possible lexical semantic interpretations, rather than over a set of discrete syntactic structures.
The probability distributions in \cite{levy_expectation-based_2008} can be seen as approximations of the neural activation distributions in our model, since more active representations are more likely to be selected. 
In fact, \cite{levy_expectation-based_2008} speculates that the probability distributions in his model likely arise from neural activation distributions (p. 1135, footnote 8). 

Both neural activation distributions and probability distributions are dynamic, since they change in response to linguistic input. 
A crucial difference, though, is that the neural activation distributions in our model, and not the probability distributions in \cite{levy_expectation-based_2008}, have \textit{internal} dynamics. 
They change not only in response to linguistic input, but also in response to their own internal state, via within-field and cross-field interaction. 
This makes our model more complex than that of \cite{levy_expectation-based_2008}; it is worth spelling out what justifies this additional complexity.

The first advantage of internal neural dynamics over internally-static probability distributions is that the former offers a built-in mechanism for interpretation selection: activation peak stabilization. 
No such mechanism exists in \cite{levy_expectation-based_2008}; it would have to be stipulated. 
A reasonable stipulation would be that interpretation selection is a sampling of the probability distribution at some timestep of processing.
This stipulation correctly predicts the observed effect of context on acceptability: the selected interpretation of the target sentence would be probabilistically biased by the context sentence.
Since response times are related to change in probability distribution, a further stipulation is required to predict the interaction between context, acceptability, and response time: that the selection process brings the probability of the selected interpretation to 1, and that of all other interpretations to 0.
Under this set of stipulations, the Bayesian inference model could be made to generate the same predictions as the DFT model.
The internal neural dynamics of DFT, while apparently increasing model complexity relative to a Bayesian model, obviate the need for such stipulations.

A second advantage of the internal neural dynamics of DFT is that they generate novel predictions, e.g., by relating the interaction of representations to their metric distance in feature space.
In general, interaction is excitatory for more similar representations, and inhibitory for more dissimilar representations.
These are the dynamics which allow the stabilization of activation peaks; neural recordings consistent with these dynamics have been observed in cat visual cortex, for example \citep{jancke_parametric_1999}.
With respect to language, it has been observed that different readings of homophonous words---which are more distal in semantic space---tend to inhibit each other, leading to slowed response times, while different readings of polysemous words---which are more proximal in semantic space---tend to prime each other, leading to faster response times \citep{frisson_about_2015,klepousniotou_disambiguating_2007,klepousniotou_not_2012,macgregor_sustained_2015,rodd_making_2002}.
From the current perspective, these effects can be seen as a natural consequence of the basic dynamics of lateral interaction in neural fields, analogous to effects of metric feature distance on target-distractor interaction observed in the domains of eye saccades \citep{kopecz_saccadic_1995}, manual reaching movements \citep{erlhagen_dynamic_2002}, and speech articulation \citep{tilsen_subphonemic_2009}.
Careful variation of the semantic distance between readings in priming studies could potentially be used to empirically constrain the parameters of lateral interaction which determine the radii of excitatory and inhibitory projection in neural fields governing processing on semantic dimensions.

In summary, Bayesian inference approaches like that of \cite{levy_expectation-based_2008} have an advantage of simplicity.
We propose that the additional complexity in a DFT-based model like ours is worthwhile since the model generates a wider variety of novel empirical predictions, and requires less stipulation to link model behavior to human behavior.
More broadly, our model offers a \textit{neural process-based} account of language comprehension, which is beyond the scope of a Bayesian inference model.

\subsection{Conclusion}
\label{conclusion}

Using a single theoretical framework (DFT) and mathematics of description (differential equations) allows explicit integration across the cognitive and sensory-motor domains.
Previous DFT modeling work has linked neural representations of conceptual structure \citep{jackendoff_foundations_2002} with visual perception \citep{grieben_bridging_2022} and visual search behavior \citep{sabinasz_neural_2023}.
Explicit coupling between abstract cognitive processes and sensation/movement has been termed ``grounding" of cognition \citep[e.g.,][]{sabinasz_neural_2023-1,sabinasz_neural_2023}.
As described in Section \ref{sec:DFT_language}, existing DFT models of speech and language have focused on phonetic dimensions of articulatory movement and auditory perception, or perceptual dimensions of objects like color, size, and spatial position.
In this paper we have applied DFT to model linguistic semantics through the lens of lexical polysemy, a cognitive domain that appears relatively distinct from sensation and behavior.
Our use of the same theoretical framework and mathematics as these previous models paves the way for explicitly coupling the cognitive and sensory-motor aspects of language.
This would represent a significant step towards a more integrated neurocognitive model of language linking meaning and form: a grounded model of linguistic cognition.

\section{Data availability}
\label{sec:OSF}
Data (experimental and simulated) and scripts for analysis and simulation are available on OSF at \cite{stern_contextual_data}.

\section{Funding sources}
\label{funding}
This work was supported by a research award from Yale University to the Neurolinguistics Lab.

\appendix
\section{Model equations and descriptions}

The dynamics of the \textit{have} node are given in Eq. \ref{eq:node}.
\label{appendix:equations}
\begin{equation}\label{eq:node}
\tau\dot{u}(t) = -u(t) + s_{\text{ext}}(t) + q\xi(t)
\end{equation}
The rate of change of activation $\dot{u}(t)$ is negatively related to current activation $u(t)$, defining a dynamical system with a point attractor at $s_{ext}(t)+q\xi(t)$. 
$s_{\text{ext}}(t)$ represents external input to the node, e.g. from perception or intention, and $\xi(t)$ represents normally distributed noise weighted by a parameter $q$. 
When there is external input $s_{ext}(t)$ to the node, the node’s activation is attracted to $s_{ext}(t)$ $(+\ q\xi(t))$. 
When there is no input, activation is attracted to 0 $(+\ q\xi(t))$.
$\tau$ is a time constant, with higher values corresponding to slower rates of evolution. 
In our simulations, $\tau$ is set to $5$, and $q$ is set to $1$.
We set the magnitude of external input $s_{\text{ext}}$ depending on the condition being simulated. 

The dynamics of each of the two DNFs are given in Eq. \ref{eq:DNF}.
\begin{equation}\label{eq:DNF}
\begin{split}
\tau \dot{u}(x,t) = -u(x,t) + h + s_{\text{ext}}(x,t) + s_{\text{node}}(x,t) + s_{\text{DNF}}(x,t)\\
+ \int k(x-x')g(u(x',t))dx' + q\xi(x,t)
\end{split}
\end{equation}
Activation $u$ is defined for each neuron $x$ representing the relevant semantic dimension at each moment in time $t$. 
Activation in the DNF has a point attractor at $h+s_{\text{ext}}(x,t) + s_{\text{node}}(x,t) + s_{\text{DNF}}(x,t) + \int k(x-x')g(u(x',t))dx' + q\xi(x,t)$. 
The resting level $h$ is assumed to be below zero for all neurons, by convention at $-5$. 
We set the time constant $\tau$ for the DNFs to $20$.

Each input $s_{\text{ext}}(x,t)$, $s_{\text{node}}(x,t)$, and $s_{\text{DNF}}(x,t)$ is represented as a separate Gaussian distribution of the form 
\begin{equation}\label{eq:input}
s(x,t) = a \exp\left[-\frac{(x-p)^2}{2w^2}\right]
\end{equation}
where $a$ controls the amplitude of the input, $p$ controls the position of the input in the field, and $w$ controls the width of the input distribution. 
In our simulations, we set the amplitude $a_{\text{ext}}$ of external input depending on the condition being simulated. 
$s_{\text{node}}(x,t)$ represents input from the \textit{have} node, whose amplitude is defined straightforwardly as a linear function of the activation of the node: $a_{\text{node}}(t) = u(node,t)$. 
This is a simplification relative to most DFT models, where the amplitude of input from a node to a field would be a more complex (sigmoidal) function of node activation.
Our motivation for eliminating the sigmoidal gating function on node-to-field input comes from work on lexical neighborhood effects on articulation, where non-selected lexical items (nodes) exert some influence on phonetic planning (DNFs) \citep{stern_not_2023}.
This issue is largely orthogonal to the present study because there is only one node in the model.
We chose to use the same node-field coupling dynamics as in \cite{stern_not_2023} for the sake of simplicity and consistency.
The same results could likely be achieved with a relatively shallow or ``soft" sigmoidal gating function.

The amplitude $a_{\text{DNF}}$ of each input from one DNF to the other is given by
\begin{equation}\label{eq:DNF_coupling}
a_{DNF} = \frac{{\max(u(\text{sender})) - \max(u(\text{receiver}))}}{{\max(u(\text{sender})) - h}} \cdot \sum_{i}^{F_{\text{sender}}} \frac{c_{DNF} \cdot (u(x_i) - h)}{1 + (\frac{x_i - p_{\text{sender}}}{w_{\text{sender}}})^4}
\end{equation}

The term on the right, $\sum_{i}^{F_{\text{sender}}} \frac{c_{DNF} \cdot (u(x_i) - h)}{1 + (\frac{x_i - p_{\text{sender}}}{w_{\text{sender}}})^4}$, defines the basic magnitude of $a_{\text{DNF}}$. 
The magnitude of input from each neuron $x_i$ in the field which is sending input (of size $F_{\text{sender}}$) to the field receiving input is determined by its activation above resting level $h$, weighted by a parameter $c_{\text{DNF}}$, set to $0.35$ in our simulations. 
This magnitude is further weighted by $1 + (\frac{x_i - p_{\text{sender}}}{w_{\text{sender}}})^4$, a nonlinear (quartic) function of the distance from $x_i$ to the center of the sending distribution $p_{\text{sender}}$, divided by the width of the sending distribution $w_{\text{sender}}$. 
Neurons within one $w_{\text{sender}}$ of $p_{\text{sender}}$ contribute substantially to $a_{\text{DNF}}$, while neurons exceeding one $w_{\text{sender}}$ from $p_{\text{sender}}$ contribute exponentially less. 
This value is summed for all neurons within the sending field, and then weighted by the term on the left, $\frac{{\max(u(\text{sender})) - \max(u(\text{receiver}))}}{{\max(u(\text{sender})) - h}}$. 
When the maximum activation within the sending distribution (ranging from $p_{\text{sender}} - w_{\text{sender}}$ to $p_{\text{sender}} + w_{\text{sender}}$) is much greater than the maximum activation in the receiving distribution, this term approaches 1, so $a_{\text{DNF}} \approx \sum_{i}^{F_{\text{sender}}} \frac{c_{DNF} \cdot (u(x_i) - h)}{1 + (\frac{x_i - p_{\text{sender}}}{w_{\text{sender}}})^4}$. 
However, as the maximum activation in the receiving distribution approaches (or exceeds) the maximum activation in the sending distribution, the weighting term approaches 0, reducing or eliminating $s_{\text{DNF}}(x,t)$. 
In this way, the sending field cannot increase activation in the receiving field beyond its own maximum activation, preventing an infinite positive feedback loop.
The centers $p$ and widths $w$ of each distribution are given in Table \ref{tab:coupling_parameters}.

\begin{table}
    \centering
    \begin{tabular}{c c c}
    \hline
    \thead{distribution} & \thead{$p$} & \thead{$w$}\\
         \hline
         \thead{low CA} & $30$ & $20$ \\
         \thead{high CA} & $70$ & $20$ \\
         \thead{low conn} & $25$ & $12$ \\
         \thead{mid conn} & $50$ & $12$ \\
         \thead{high conn} & $75$ & $12$ \\
         \hline
    \end{tabular}
    \caption{DNF coupling distributions.}
    \label{tab:coupling_parameters}
\end{table}
	
\begin{figure}[h]
    \centering
    \includegraphics[width=1\linewidth]{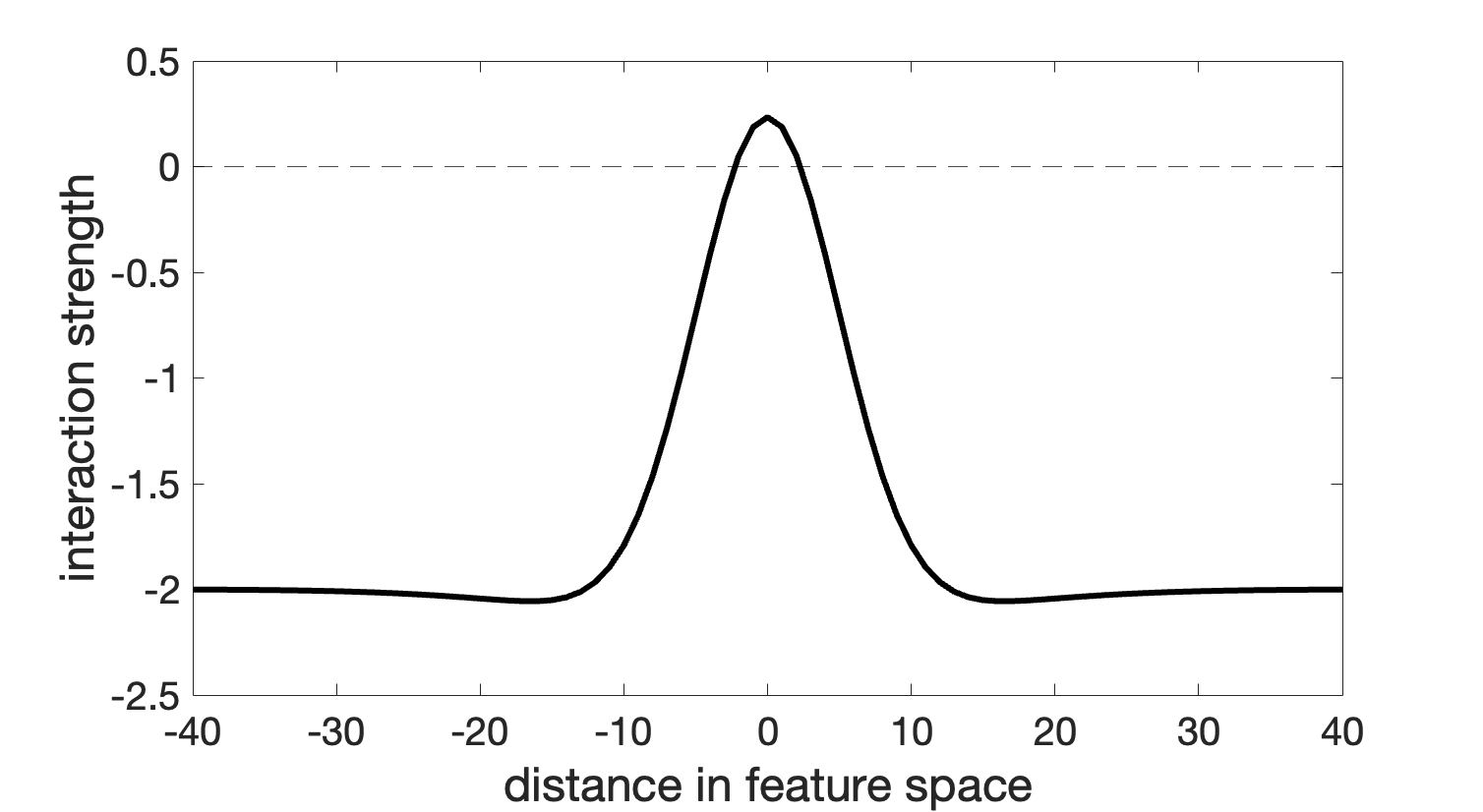}
    \caption{Lateral interaction kernel $k(x-x')$.}
    \label{fig:lateral_interaction}
\end{figure}

Within-field lateral interaction between neurons is defined by an interaction kernel $k(x-x')$: 

\begin{equation}\label{eq:lateral_interaction}
\begin{split}
k(x-x') = \frac{c_{exc}}{\sqrt{2\pi} \sigma_{exc}} \exp\left[-\frac{(x-x')^2}{2\sigma_{exc}^2}\right] \\
- \frac{c_{inh}}{\sqrt{2\pi} \sigma_{inh}} \exp\left[-\frac{(x-x')^2}{2\sigma_{inh}^2}\right] - c_{glob}
\end{split}
\end{equation}
Each neuron $x'$ which exceeds an activation threshold contributes activation to other neurons $x$ as a function of their distance within the field $(x-x')$. 
As seen in Figure \ref{fig:lateral_interaction}, interaction is excitatory (weighted by $c_{exc}$, set here to $30$) for nearby neurons (defined by $\sigma_{exc}$, set here to $5$) and inhibitory (weighted by $c_{inh}$, set here to $5$) for more distant neurons (defined by $\sigma_{inh}$, set here to $12.5$). 
$c_{glob}$, set here to 2, contributes global inhibition from each above-threshold neuron.
Lateral excitation helps to stabilize activation peaks (which correspond to semantic interpretations), and lateral inhibition prevents runaway expansion of activation peaks. 
Crucially, we set the parameters of the interaction kernel such that only a single peak can form at a time in a given field for the range of input amplitudes under consideration in our simulations, defining selection dynamics. 

\begin{figure}[h]
    \centering
    \includegraphics[width=1\linewidth]{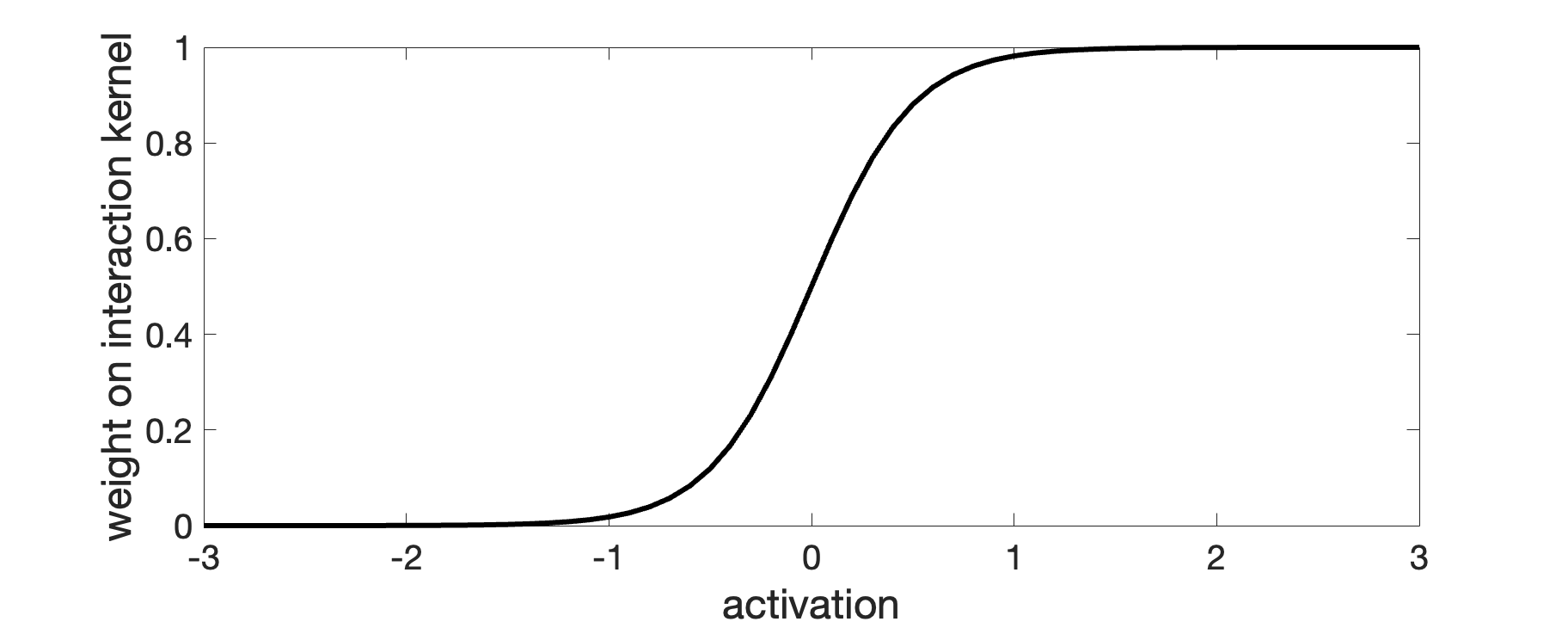}
    \caption{Sigmoidal function $g(u)$ gating lateral interaction.}
    \label{fig:sigmoid}
\end{figure}

As seen in Eq. \ref{eq:sigmoid} and Figure \ref{fig:sigmoid}, the activation threshold for interaction is given by a sigmoidal function $g(u)$, where $\beta$ (set here to $4$) controls the steepness of the threshold:
\begin{equation}\label{eq:sigmoid}
g(u) = \frac{1}{{1+\exp(-\beta u)}}
\end{equation}
By convention, the threshold is set to $u = 0$ so that lateral interaction kicks in only when activation approaches $0$.
Finally, noise in field activation is simulated by adding normally distributed random values $\xi(x,t)$ weighted by $q$, set here to $4$. 

\section{Experimental stimuli}
\label{appendix:stimuli}
\begin{longtable}{|l|l|l|p{0.35\textwidth}|}
\hline
        \textbf{stimulus set} & \textbf{stimulus type} & \textbf{condition} & \textbf{stimulus} \\ \hline
        \endfirsthead
        \hline
        \textbf{stimulus set} & \textbf{stimulus type} & \textbf{condition} & \textbf{stimulus} \\ \hline
        \endhead
        \hline
        \endfoot
        \hline
        \endlastfoot
        1 & experimental & adjacency & The maple tree has a plastic box behind it and the oak tree has a skateboard that is red. \\ \hline
        1 & experimental & possession & The maple tree has a branch that is thick and the oak tree has a skateboard that is red. \\ \hline
        2 & experimental & adjacency & The streetlight has a pink scooter next to it and the traffic sign has a shrub that is short. \\ \hline
        2 & experimental & possession & The streetlight has a bulb that is bright and the traffic sign has a shrub that is short. \\ \hline
        3 & experimental & adjacency & The wooden desk has a small bottle inside of it and the metal desk has a pen that is old. \\ \hline
        3 & experimental & possession & The wooden desk has a drawer that is wide and the metal desk has a pen that is old. \\ \hline
        4 & experimental & adjacency & The brick building has a new bus behind it and the concrete building has a crate that is large. \\ \hline
        4 & experimental & possession & The brick building has a window that is broken and the concrete building has a crate that is large. \\ \hline
        5 & experimental & adjacency & The table has a large plate on top of it and the chair has a book that is brown. \\ \hline
        5 & experimental & possession & The table has a leg that is wobbly and the chair has a book that is brown. \\ \hline
        6 & experimental & adjacency & The blanket has a soft pillow underneath it and the sheet has a bottle that is plastic. \\ \hline
        6 & experimental & possession & The blanket has a pattern that is beautiful and the sheet has a bottle that is plastic. \\ \hline
        7 & experimental & adjacency & The textbook has a small mug next to it and the notebook has a stapler that is silver. \\ \hline
        7 & experimental & possession & The textbook has a page that is torn and the notebook has a stapler that is silver. \\ \hline
        8 & experimental & adjacency & The skillet has a yellow onion next to it and the pan has a glass that is tall. \\ \hline
        8 & experimental & possession & The skillet has a handle that is sturdy and the pan has a glass that is tall. \\ \hline
        9 & experimental & adjacency & The door has a green fern next to it and the mirror has a painting that is large. \\ \hline
        9 & experimental & possession & The door has a lock that is strong and the mirror has a painting that is large. \\ \hline
        10 & experimental & adjacency & The monitor has a wooden ruler behind it and the laptop has a cup that is black. \\ \hline
        10 & experimental & possession & The monitor has a screen that is spotless and the laptop has a cup that is black. \\ \hline
        11 & filler-related & adjacency & The maple tree conceals a plastic box behind it and the oak tree has a skateboard that is red. \\ \hline
        11 & filler-related & possession & The maple tree possesses a branch that is thick and the oak tree has a skateboard that is red. \\ \hline
        12 & filler-related & adjacency & The streetlight abuts a scooter that is pink and the traffic sign has a shrub that is short. \\ \hline
        12 & filler-related & possession & The streetlight possesses a bulb that is bright and the traffic sign has a shrub that is short. \\ \hline
        13 & filler-related & adjacency & The wooden desk holds a small bottle inside of it and the metal desk has a pen that is old. \\ \hline
        13 & filler-related & possession & The wooden desk possesses a drawer that is wide and the metal desk has a pen that is old. \\ \hline
        14 & filler-related & adjacency & The brick building conceals a new bus behind it and the concrete building has a crate that is large. \\ \hline
        14 & filler-related & possession & The brick building possesses a window that is broken and the concrete building has a crate that is large. \\ \hline
        15 & filler-related & adjacency & The table holds a large plate on top of it and the chair has a book that is brown. \\ \hline
        15 & filler-related & possession & The table possesses a leg that is wobbly and the chair has a book that is brown. \\ \hline
        16 & filler-related & adjacency & The blanket covers a pillow that is soft and the sheet has a bottle that is plastic. \\ \hline
        16 & filler-related & possession & The blanket possesses a pattern that is beautiful and the sheet has a bottle that is plastic. \\ \hline
        17 & filler-related & adjacency & The textbook abuts a mug that is small and the notebook has a stapler that is silver. \\ \hline
        17 & filler-related & possession & The textbook contains a page that is torn and the notebook has a stapler that is silver. \\ \hline
        18 & filler-related & adjacency & The skillet abuts an onion that is yellow and the pan has a glass that is tall. \\ \hline
        18 & filler-related & possession & The skillet possesses a handle that is sturdy and the pan has a glass that is tall. \\ \hline
        19 & filler-related & adjacency & The door abuts a fern that is green and the mirror has a painting that is large. \\ \hline
        19 & filler-related & possession & The door possesses a lock that is strong and the mirror has a painting that is large. \\ \hline
        20 & filler-related & adjacency & The monitor conceals a wooden ruler behind it and the laptop has a cup that is black. \\ \hline
        20 & filler-related & possession & The monitor possesses a screen that is spotless and the laptop has a cup that is black. \\ \hline
        21 & filler-unrelated & metonymy & A hotel guest approaches the hotel's front desk and asks the clerk: “Has room 451 checked out yet?” \\ \hline
        21 & filler-unrelated & no metonymy & A hotel guest approaches the hotel's front desk and asks the clerk: “Has room 451 been cleaned yet?” \\ \hline
        22 & filler-unrelated & metonymy & In a hospital, a medical assistant asks the doctor: “Bed 22's granddaughter is here to see him. Can she go in?” \\ \hline
        22 & filler-unrelated & no metonymy & In a hospital, a medical assistant asks the doctor: “Ed Montague’s granddaughter is here to see him. Can she go in?” \\ \hline
        23 & filler-unrelated & metonymy & One waiter in a busy restaurant complains to another: “You forgot to tell the chardonnay at table 6 about the specials!” \\ \hline
        23 & filler-unrelated & no metonymy & One waiter in a busy restaurant complains to another: “You forgot to tell the customer at table 6 about the specials!” \\ \hline
        24 & filler-unrelated & metonymy & In a diner, one waitress tells another: “The ham sandwich in the corner needs another cup of coffee.” \\ \hline
        24 & filler-unrelated & no metonymy & In a diner, one waitress tells another: “The tall woman in the corner needs another cup of coffee.” \\ \hline
        25 & filler-unrelated & metonymy & At an art shop, the owner tells the frame maker: “The Monet will pick up his order on Friday.” \\ \hline
        25 & filler-unrelated & no metonymy & At an art shop, the owner tells the frame maker: “The professor will pick up his order on Friday.” \\ \hline
        26 & filler-unrelated & metonymy & The server in a sandwich shop tells the chef: “The tuna on rye wants extra mayonnaise and a side salad.” \\ \hline
        26 & filler-unrelated & no metonymy & The server in a sandwich shop tells the chef: “The tuna on rye comes with mayonnaise and a side salad.” \\ \hline
        27 & filler-unrelated & metonymy & One stylist in a hair salon says to another: “The highlights sitting in chair 3 needs a blow-dry.” \\ \hline
        27 & filler-unrelated & no metonymy & One stylist in a hair salon says to another: “The lady sitting in chair 3 needs a blow-dry.” \\ \hline
        28 & filler-unrelated & metonymy & In a crowded emergency room, a nurse tells the doctor: “Room 3's blood pressure is very high, and he feels dizzy.” \\ \hline
        28 & filler-unrelated & no metonymy & In a crowded emergency room, a nurse tells the doctor: “Mr. Lee’s blood pressure is very high, and he feels dizzy.” \\ \hline
        29 & filler-unrelated & metonymy & In an expensive hotel, the front desk clerk tells the chef: “Room 12 ordered room service, they want the lobster.” \\ \hline
        29 & filler-unrelated & no metonymy & In an expensive hotel, the front desk clerk tells the chef: “Mr. Dell ordered room service, they want the lobster.” \\ \hline
        30 & filler-unrelated & metonymy & A waiter in a burger joint yells to the cook: “The cheeseburger at table 10 just asked for extra pickles!” \\ \hline
        30 &  filler-unrelated & no metonymy & A waiter in a burger joint yells to the cook: “The cheeseburger for table 10 should come with extra pickles!” \\ \hline
        31 & filler-unrelated & metonymy & On an airplane, a flight attendant tells the pilot: “Seat 25D will not sit down, so we can't take off yet.” \\ \hline
        31 & filler-unrelated & no metonymy & On an airplane, a flight attendant tells the pilot: “One passenger will not sit down, so we can't take off yet.” \\ \hline
        32 & filler-unrelated & metonymy & In a seafood restaurant, a waiter tells the busboy: “The clam chowder at table 3 ordered a glass of wine.” \\ \hline
        32 & filler-unrelated & no metonymy & In a seafood restaurant, a waiter tells the busboy: “The blonde lady at table 3 ordered a glass of wine.” \\ \hline
        33 & filler-unrelated & metonymy & In a university, a professor tells her students: “Classroom 217 is still in session, so we'll have to wait.” \\ \hline
        33 & filler-unrelated & no metonymy & In a university, a professor tells her students: “Classroom 217 is still being used, so we'll have to wait.” \\ \hline
        34 & filler-unrelated & metonymy & In a steak house downtown, a waitress calls to the chef: “The filet mignon wants it with rice and vegetables!” \\ \hline
        34 & filler-unrelated & no metonymy & In a steak house downtown, a waitress calls to the chef: “The filet mignon comes with rice and vegetables!” \\ \hline
        35 & filler-unrelated & metonymy & One emergency room nurse says to another: “The appendicitis in room 17B says she needs another pain pill.” \\ \hline
        35 & filler-unrelated & no metonymy & One emergency room nurse says to another: “The patient in room 17B says she needs another pain pill.” \\ \hline
        36 & filler-unrelated & metonymy & A frazzled waiter forgets the orders, so when he brings the food he asks Sue: “Are you the pasta or the steak?” \\ \hline
        36 & filler-unrelated & no metonymy & A frazzled waiter forgets the orders, so when he brings the food he asks Sue: “Is yours the pasta or the steak?” \\ \hline
        37 & filler-unrelated & metonymy & Just before the concert started, the conductor tells the orchestra: “The violins have the flu and can't play tonight.” \\ \hline
        37 & filler-unrelated & no metonymy & Just before the concert started, the conductor tells the orchestra: “The soloists have the flu and can't play tonight.” \\ \hline
        38 & filler-unrelated & metonymy & A barista at Starbucks this morning forgets my coffee order, then he asks me: “Are you the latte or the cappuccino?” \\ \hline
        38 & filler-unrelated & no metonymy & A barista at Starbucks this morning forgets my coffee order, then he asks me: “Is yours the latte or the cappuccino?” \\ \hline
        39 & filler-unrelated & metonymy & In an expensive French restaurant, one waiter says to another: “Table 4 asked for another bottle of chardonnay.” \\ \hline
        39 & filler-unrelated & no metonymy & In an expensive French restaurant, one waiter says to another: “That couple asked for another bottle of chardonnay.” \\ \hline
        40 & filler-unrelated & metonymy & One bartender in a cocktail lounge says to another: “The mojito at table 6 has been flirting with you all night.” \\ \hline
        40 & filler-unrelated & no metonymy & One bartender in a cocktail lounge says to another: “The girl at table 6 has been flirting with you all night.” \\ \hline
        41 & practice & grammatical & It is a beautiful day so the family is relaxing outside \\ \hline
        42 & practice & grammatical & The math test is tomorrow so the student is studying. \\ \hline
        43 & practice & ungrammatical & The game is tomorrow so the player practicing is after school. \\ \hline
        44 & practice & ungrammatical & There is a new arcade so Bill play video games every day. 
    \label{tab:stimuli}
\end{longtable}

\bibliographystyle{elsarticle-harv} 
\bibliography{cas-refs}

\end{document}

%% file: titlepage.tex
\begin{titlepage}

\title{Contextual modulation of language comprehension in a dynamic neural model of lexical meaning}

\author[inst1]{Michael C. Stern}

\affiliation[inst1]{organization={Linguistics Department, Yale University},
            addressline={370 Temple St.}, 
            city={New Haven},
            state={Connecticut},
            postcode={06511},
            country={USA}}
\ead{michael.stern@yale.edu}

\author[inst1]{Maria M. Piñango}

\end{titlepage}